\documentclass[nohyperref]{article}

\usepackage{microtype}
\usepackage{graphicx}
\usepackage{subfigure}
\usepackage{booktabs} %

\usepackage{xspace}

\usepackage{hyperref}

\usepackage[accepted]{icml2022_modified}  %

\usepackage{amsmath}
\usepackage{amssymb}
\usepackage{mathtools}
\usepackage{amsthm}

\usepackage[capitalize,noabbrev]{cleveref}

\theoremstyle{plain}

\theoremstyle{definition}

\theoremstyle{remark}

\icmltitlerunning{StreamingQA}

\begin{document}

\twocolumn[

\icmltitle{StreamingQA: A Benchmark for Adaptation to New Knowledge over Time\\ in Question Answering Models}

\icmlsetsymbol{equal}{*}
\icmlsetsymbol{lead}{$\spadesuit$}

\begin{icmlauthorlist}
\icmlauthor{Adam Li\v{s}ka}{equal,gl}
\icmlauthor{Tom\'{a}\v{s} Ko\v{c}isk\'{y}}{equal,lead,dm}
\icmlauthor{Elena Gribovskaya}{equal,lead,dm}
\icmlauthor{Tayfun Terzi}{equal,dm}\\
\icmlauthor{Eren Sezener}{dm}
\icmlauthor{Devang Agrawal}{gl}
\icmlauthor{Cyprien de Masson d'Autume}{dm}
\icmlauthor{Tim Scholtes}{dm}
\icmlauthor{Manzil Zaheer}{dm}
\icmlauthor{Susannah Young}{dm}
\icmlauthor{Ellen Gilsenan-McMahon}{dm}
\icmlauthor{Sophia Austin}{dm}
\icmlauthor{Phil Blunsom}{co,csox}
\icmlauthor{Angeliki Lazaridou}{dm}
\end{icmlauthorlist}

\icmlaffiliation{dm}{DeepMind, London, UK}
\icmlaffiliation{csox}{Department of Computer Science, University of Oxford, Oxford, UK}
\icmlaffiliation{gl}{Glyphic~AI, work done at DeepMind}
\icmlaffiliation{co}{Cohere, work done at DeepMind}

\icmlcorrespondingauthor{Tom\'{a}\v{s} Ko\v{c}isk\'{y}}{tkocisky@deepmind.com}
\icmlcorrespondingauthor{Elena Gribovskaya}{egribovskaya@deepmind.com}

\icmlkeywords{Machine Learning, ICML}

\vskip 0.3in
]

\renewcommand{\icmlEqualContribution}{\textsuperscript{*}Equal contribution in random order }
\printAffiliationsAndNotice{
\icmlEqualContribution
$^\spadesuit$Project Leads 
} %

\setlength{\textfloatsep}{\textfloatsep-2mm}

\newcommand{\streamingqa}{StreamingQA\xspace}  %

\begin{abstract}

Knowledge and language understanding of models evaluated  through question  answering (QA)  has  been  usually  studied  on  static  snapshots of knowledge, like Wikipedia. However, our world is dynamic, evolves over time, and our models' knowledge becomes outdated. To study how semi-parametric QA models and their underlying parametric language models (LMs) adapt to evolving knowledge, we construct a new large-scale dataset, \streamingqa, with human written and generated questions  asked  on  a  given  date, to be answered from 14 years of time-stamped news articles. We evaluate our models quarterly as they read new articles not seen in pre-training. We show that parametric models can be updated without full retraining, while avoiding catastrophic forgetting. For semi-parametric models, adding new articles into the search space allows for rapid adaptation, however, models with an outdated underlying LM under-perform those with a retrained LM. For questions about higher-frequency named entities, parametric updates are particularly beneficial. In our dynamic world, the \streamingqa dataset enables a more realistic evaluation of QA models, and our experiments highlight several promising directions for future research.

\end{abstract}

\section{Introduction}
\label{sec:intro}

\begin{table}[t]
    \caption{Example questions from \streamingqa\ (Eval-Written).
    In addition to a question date, each question has three reference answers, and an evidence document with its publication date.
    }
    \label{tab:examples_two}
    \small
    \centering
    \begin{tabular}{p{0.85\linewidth}}
    \toprule
    \textit{Recent subset} \\
    \underline{Question Date:} Monday, February 24, 2020 \\
    \underline{Question:} How many countries have committed to the net zero target as of today's date? \\
    \midrule
    \textit{Past subset} \\
    \underline{Question Date:} Sunday, April 12, 2020 \\
    \underline{Question:} In November 2016, which Netflix series set in the United Kingdom was said to be ``the most expensive television series ever''?\\
    \bottomrule
    \end{tabular}
\end{table}

Question answering (QA) allows us to interrogate models for their language understanding, knowledge, and reasoning abilities, while also being useful in various knowledge-oriented applications such as personal assistants or web search.
The questions that people ask span all our knowledge and can be about any point in the history, although often they are about the most recent events that happened in the last few weeks or days. 
Consider examples in Table~\ref{tab:examples_two} that ask about events as distant as 4 years before or as recent as the day when the question was asked.
As the world and knowledge evolve, we need our QA models to adapt to new information, to not forget the past, and to maintain an up-to-date world model to make our interaction with such systems more meaningful. To evaluate and improve models' ability to adapt, we need a dataset with temporal grounding of both questions and knowledge---dates when questions were asked and publication dates of documents.
As the currently available QA datasets are not suitable for this, we propose a novel dataset and subsequently perform a systematic study of adaptation in the state-of-the-art QA models.

Previous research has often focused on answering questions about individual passages or books \cite{rajpurkar-etal-2016-squad,kocisky-etal-2018-narrativeqa}, answering about static structured \cite{timequestions2021}, or unstructured knowledge corpora such as Wikipedia \cite{lee-etal-2019-latent-NQ-OPEN}.
More recent work has considered answering questions about knowledge with temporal grounding of facts in a knowledge graph \cite{saxena-etal-2021-question, timequestions2021} or news articles \cite{wang2021archivalqa, dhingra2021timeaware} but without grounding the questions with a question date;
or have repurposed questions from other datasets and added question dates
\cite{zhang-choi-2021-situatedqa} but still answered over a static snapshot of Wikipedia.

We present a new dataset, \streamingqa\footnote{\url{https://github.com/deepmind/streamingqa}}, that provides temporal context of both, the questions and the knowledge required to answer them. 
The dataset contains questions written by annotators or generated with a large-scale LM. The questions are answerable from a \emph{streaming} knowledge corpus of time-stamped English WMT
news articles published between 2007 and 2020 (see Figure~\ref{fig:dataset}).
Having temporal metadata for questions and articles enables us to ingest new knowledge periodically,
in a streaming setup,
and evaluate on questions asked during that period. We consider questions about recent and past knowledge separately to measure adaptation and forgetting.
Moreover, question dates allow to ask questions with relative time specifications (e.g., ``3 months ago''), which are under-represented in the existing QA datasets.
Lastly, news domain, compared to often used Wikipedia, provides more realistic challenges for open-book retrieval with redundant, noisy, and sometimes conflicting information. 

Previous work demonstrated that large LMs struggle with temporal generalization, a type of domain shift that occurs when a model at test time needs to understand new knowledge, named entities, and topics \cite{lazaridou2021mind, rottger2021temporal}. In this work, we leverage \streamingqa to quantify similar adaptation 
in existing parametric (closed-book) and semi-parametric (open-book) QA models that today are frequently based on such LMs. 

Our findings suggest that parametric adaptation improves QA performance for open-book approaches
like RAG \cite{Lewis2020RetrievalAugmentedGF} and FiD \cite{izacard2020leveraging} (Section~\ref{sec:results_ob}). Moreover, a more granular, frequency-based analysis (Section~\ref{sec:results_param_vs_semiparam}) suggests that parametric and semi-parametric approaches to adaptation are complementary: parametric adaptation improves accuracy of questions about frequent knowledge/named entities, where semi-parametric, dense retrieval-based methods can under-perform \cite{DBLP:journals/corr/abs-2109-01156}. In contrast, semi-parametric adaptation helps with less frequent knowledge where retrieval is less confused by redundancy and ambiguity of information associated with more frequent names, where the parametric LMs struggle.
In the closed-book setup, we find that incremental fine-tuning works reasonably well without causing catastrophic forgetting (Section~\ref{sec:results_cb}), and it also results in substantially lower computational costs than full model retraining%
\footnote{In addition to new knowledge being created incrementally over time, our interest in iterative fine-tuning is also due to
the large computational cost savings. In our setup, fine-tuning on a single month requires 95\% fewer steps than full retraining.}.
Lastly, we also establish benchmarks for less computationally intensive QA tasks (Section~\ref{sec:results_static}): one-step adaptation and the usual static open-book QA task.

\section{\streamingqa\ Dataset and Task}
\label{sec:streamingqa}

\begin{table*}[t]
    \caption{Examples of written and generated questions from the recent and past subsets. %
    }
    \label{tab:examples}
    \centering
\tiny
\setlength{\tabcolsep}{2pt}
    \begin{tabular}{cp{0.35\linewidth}lclll}
    \toprule
    Question Date & Question & Answer & Passage Date & Subset & Added Time & Gen./Wri. \\
    \midrule
    30.05.2020&What does Donald Trump, US president, call his 2020 plan to expedite the development of a COVID-19 vaccine?&Operation Warp Speed&19.05.2020&Recent&-&Written\\
    
    18.04.2020&In the UK, as of Sunday, April 12, 2020, how much money in loans had been given to firms seeking cash to survive the coronavirus crisis?&£800 million&12.04.2020&Recent&-&Written\\
    
    28.04.2020&Which soft drinks company was 2010 Australian  favourite racing driver Mark Webber's team sponsored by?&Red Bull&29.03.2010&Past&-&Written\\
    
    26.03.2020&Where did the research that claimed up to 50 per cent of the UK population may have already contracted the coronavirus, take place?&University of Oxford&25.03.2020&Recent&-&Generated \\
    
    11.09.2020&Which player scored for St Mirren in November 2008?&Franco Miranda&16.11.2008&Past&Absolute&Generated\\
    
    08.05.2020&Which hospital is Jamie Cooper receiving treatment at 13 years ago?&Selly Oak Hospital&01.11.2007&Past&Relative&Generated\\
    \bottomrule
\end{tabular}
\end{table*}

\newcommand{\myemph}[1]{\mbox{\emph{#1}}}

\begin{figure}[t]
    \centering
    \includegraphics[width=\linewidth]{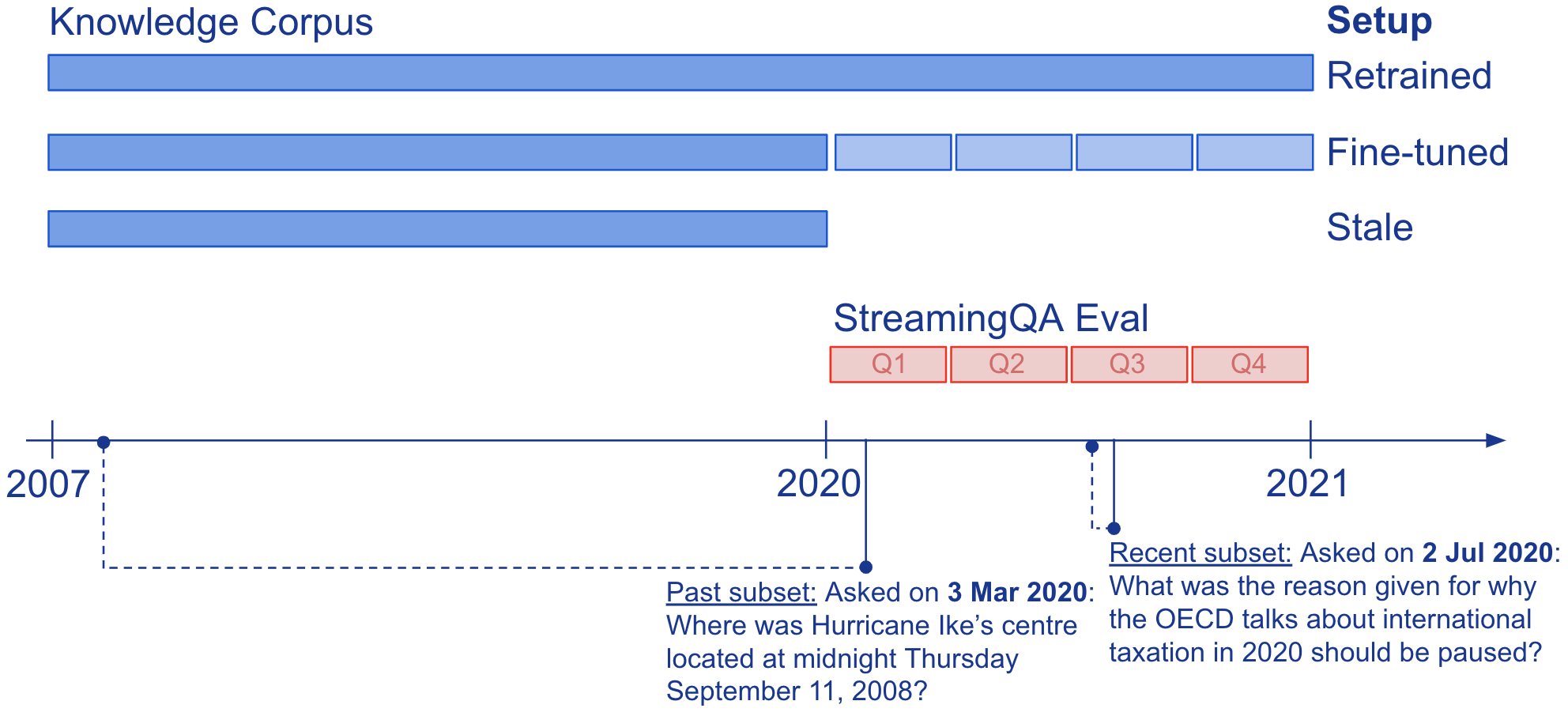}
    \vspace{-6mm}
    \caption{The \streamingqa task: we emulate a realistic scenario where a QA system needs to respond to user questions about a mix of recent and past events.
    }
    \label{fig:dataset}
\end{figure}

In this section, we introduce a new QA dataset and a task to evaluate models' adaptation and forgetting of knowledge.
We require, in addition to questions%
, temporal metadata: a question date (when the question could have been asked) and a knowledge date (when an article that answers this question was published). Using this metadata enables us to evaluate how well the model understands new knowledge that becomes available incrementally at evaluation time (see Figure~\ref{fig:dataset}). We also use the timestamps to split the training and evaluation sets into non-overlapping historical periods. 
See Table~\ref{tab:examples} for examples of questions.

To construct the \streamingqa\ dataset, we consider 14 years (2007--2020) of English WMT news articles\footnote{\url{http://data.statmt.org/news-crawl/README}}  \cite{akhbardeh2021wmt}, together with their publication dates, as our knowledge corpus (approx.\ 11M articles). 
Specifically, given an article, we first generate a question date, that is, the date when we want the question to be asked (Section~\ref{sec:qd}), and subsequently, either (i) automatically generate questions (Section~\ref{sec:gen}), or (ii) ask human annotators to write questions (Section~\ref{sec:wri}). Lastly, to reduce noise, we apply automatic and human filtering to the questions, and collect additional reference answers (Section~\ref{sec:filtering}).
We present additional statistics in Section~\ref{sec:dataset_stats}.

We consider a streaming task (Section~\ref{sec:dataset_task}): we split questions into four quarterly sets over 2020 based on their question dates, where questions in each quarter are to be answered from articles published up to and including that quarter. 
For the adaptation and forgetting analysis, we further split the quarterly evaluation sets into the \myemph{recent subset} and the \myemph{past subset}. The dataset is constructed to have an approximately equal number of each. As recent questions cover 2020 and past questions cover all history uniformly, the overall distribution is biased towards the present,
as we would expect it in an actual QA system.
The recent subset contains questions about the most recent knowledge, i.e., articles published in the month before the question date; and the past subset contains questions asked during that quarter about articles published between 2007 and the question date. Furthermore, for training and validation, we use automatically generated questions asked during 2007--2019 with an analogous split between the recent and past subsets.

\subsection{\streamingqa Task: QA with Temporal Metadata}\label{sec:dataset_task}

Formally, our dataset consists of triples of question date, question, and answer, $\mathcal Q = \{(d_{q,i}, q_i, a_i)\}_{i=1}^N$,
and a knowledge corpus of publication dates and documents,
$\mathcal C = \{(d_{c,j}, c_j)\}_{j=1}^M$.
For a given time period $t=[t_s, t_e]$ (e.g., January to March 2020), we consider questions asked during that period,
${\mathcal Q_{=t} = \{(d_{q,i}, q_i, a_i)\in\mathcal Q : t_s \leq d_{q,i} \leq t_e\}}$,
about the corresponding subset of the knowledge corpus published until then,
${\mathcal C_{\leq t} = \{(d_{c,j}, c_j)\in\mathcal C : d_{c,j} \leq t_e\}}$.
To answer a question in ${\mathcal Q_{=t}}$, we generate an answer using the corresponding knowledge corpus, ${p(a_i | q_i, d_{q,i}, \mathcal C_{\leq t})}$.

\subsection{Questions Dates}\label{sec:qd}

We need to generate question dates that are plausible with respect to article dates and make sure that the dates and events are consistent.
For evaluation sets, to create recent subset questions we sample a document with a publication date
$d_{c}\sim\mathcal U [\mathrm{Dec}2019,\mathrm{Dec}2020]$ and a question date $d_{q}\sim\mathcal U[d_c, d_c+30\mathrm{days}]$. 
To create past subset questions, we sample a document with a publication date
$d_{c}\sim\mathcal U [2007,\mathrm{Dec}2020]$ and a question date $d_{q}\sim\mathcal U[\max(d_c, \mathrm{Jan}2020), \max(d_c, \mathrm{Jan}2020)+365\mathrm{days}]$. 
The articles are distributed uniformly within a month, or across all available history prior to the question date\footnote{We filter out samples with question date not in 2020.},
for the two subsets respectively.
For training and validation sets, we consider question dates in $[2007, \mathrm{Dec}2019]$ and aim for an article distribution given a question date similar to above.

\begin{figure}
    \centering
    \subfigure[Distribution of first word for written and generated questions.]{
    \includegraphics[width=0.97\linewidth]{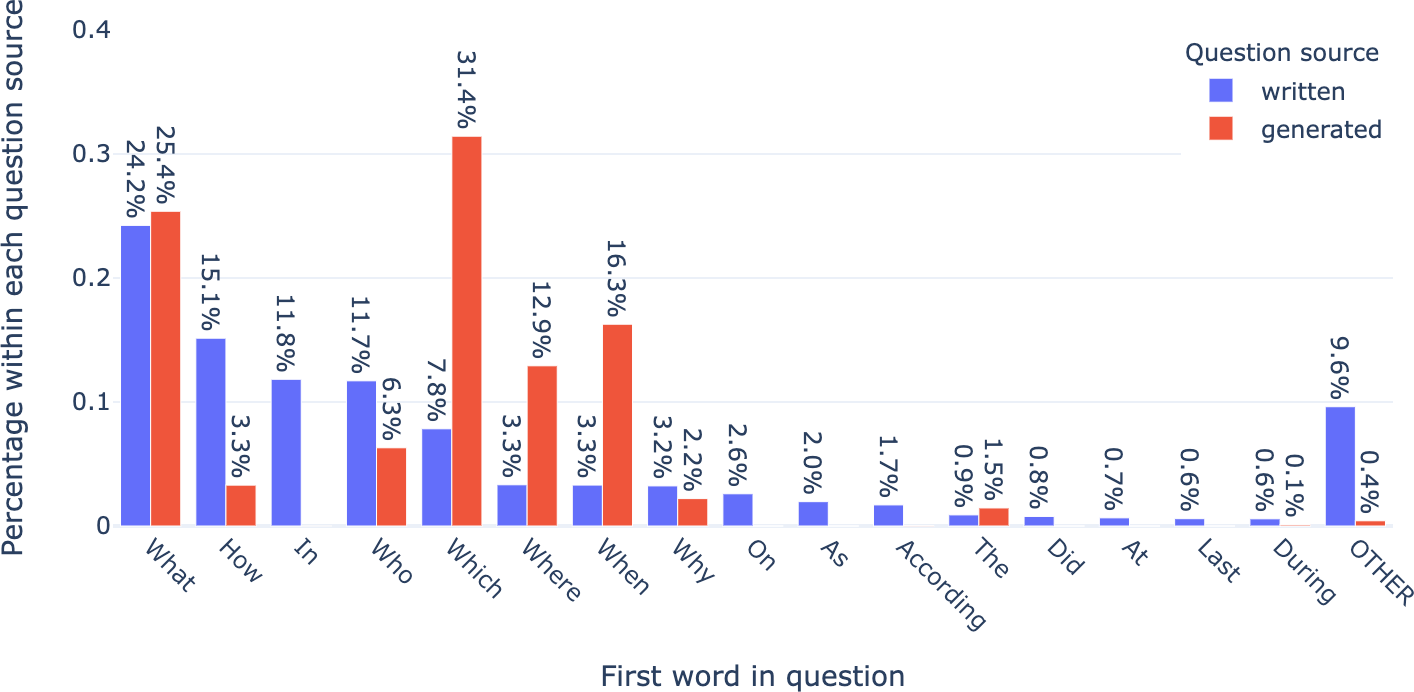}
    }
    \vspace{-2mm}
    \subfigure[Question length histograms and means.]{
    \includegraphics[width=0.47\linewidth]{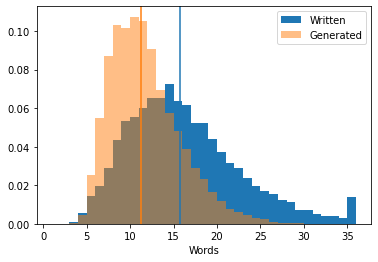}
    }
    \subfigure[Answer length histograms and means.]{
    \includegraphics[width=0.47\linewidth]{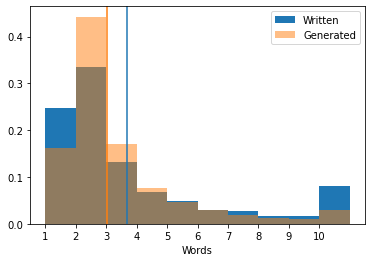}
    }
    \vspace{-2mm}
\caption{Details of the StreamingQA dataset.}
 \label{fig:streamingqa_first_word_lengths}
\end{figure}

\subsection{Automatically Generated Questions
}
\label{sec:gen}
We use automatic question generation as a scalable way to obtain questions grounded at different points in time. 
Questions are generated through few-shot prompting of a large LM \cite{rae2021scaling},
given an evidence document and a target answer drawn from named entities, dates, and prepositional phrases contained as spans in the document.
A challenge with creating questions for open-domain QA is that they need to be specific enough when considered in the context of \emph{all} articles in the knowledge corpus. We first over-generate questions-answer pairs and then apply heuristic filters to eliminate trivial and/or low quality candidates 
(Appendix~\ref{sec:app_automatic_filters}). 
For questions included in the past subset, we append absolute or relative time specifications to question text, e.g., ``3 months ago'' or ``in May 2017'' (Appendix~\ref{sec:time_spec}), unless the text already contains such a specification or the answer is a date.

\subsection{Human Written Questions}\label{sec:wri}

Human annotators were asked to write questions about a news article provided together with its publication date and desired question date. We chose annotators whose first language is English, who are in the US or the UK, and have a university education. 
Each annotator had to write up to five questions and answers about the details of the events described in the article, and framed these questions as if they were asking another person.
Each participant created about 15 questions on average. 
We explicitly asked the annotators to include enough context in the questions to make the questions as unambiguous as possible for the open-book setup.

The full details of our study design, including compensation rates, were reviewed by DeepMind’s independent ethical review committee. All participants provided informed consent prior to completing tasks and were reimbursed for their time. It is our policy that researchers must pay workers/participants at least the living wage for their location.

\subsection{Quality Filtering}\label{sec:filtering}

We filtered both generated and human written questions for quality in two stages.
First, we asked annotators to filter for good/bad question,
similar to \citet{47761-NaturalQuestions},
filtering for factual, unambiguous, grammatical questions. To judge ambiguity, we have additionally provided the question date and asked not to assume a particular location (e.g., US). To include a question, we need 3 annotators to agree.

Secondly, we asked annotators to answer each question given the original passage, its publication date, and the question date. Annotators first selected parts of the passage that supported the answer\footnote{We don't use this annotation in experiments.
}, and then wrote a short answer in their own words. We did not require the answers to be sub-strings of the passage. We only kept questions where annotators could provide answers, and obtained additional references in the same way.\footnote{We have 3 reference answers for computing the metrics, first the answer used to generate the question or provided by the original question writer, and 2 additional provided by annotators. 
}

\subsection{Statistics}\label{sec:dataset_stats}

\begin{table}[t]
    \caption{StreamingQA dataset statistics.}
    \label{tab:StreamingQAcounts}
    \centering
    \scriptsize
    \setlength{\tabcolsep}{2pt}
    \begin{tabular}{lrrrrrr}
    \toprule
              &    \multicolumn{2}{c}{Recent} &  \multicolumn{2}{c}{Past}  &  \multicolumn{2}{c}{All}     \\
              & Questions & Articles &Questions &Articles  &Questions &Articles     \\
    \midrule
    Train          &  49,451 &      49,339 &      49,951 &      49,784 &      99,402 &      98,872 \\
    Valid          &   4,966 &       4,965 &       4,973 &       4,971 &       9,939 &       9,932 \\
    Eval-Generated &  15,550 &      15,385 &      12,070 &      11,848 &      27,620 &      26,852 \\
    Eval-Written   &   4,521 &       1,545 &       4,237 &       1,448 &       8,758 &       2,993 \\
    \bottomrule
    \end{tabular}
\end{table}

The \streamingqa\ dataset contains about 28k generated questions and about 8.8k human written questions for evaluation, and 100k and 10k questions for training and validation, respectively. See Table~\ref{tab:StreamingQAcounts} for details.
In Figure~\ref{fig:streamingqa_first_word_lengths}, we see that human written questions and answers tend to be somewhat longer.
Based on the first question word distribution, we have a diverse set of both human written and generated questions, with the latter slightly biased to ``Which''/``Where''/``When'', likely due to prompting of the large LM. For written questions, many start with ``In'', which is often annotators providing temporal context so that questions stand without the article in the open-book setting. Examining the answer types on the written evaluation sets by automatic labeling, we have about 46.9\%, 40.6\%, and 12.4\% of named entity, phrase, and date answers, respectively (see Appendix~\ref{sec:app_dataset_stats}).
About 6\% of evaluation reference answers are seen among answers to questions in the training set, 
and 23\% of training questions have an answer contained in evaluation  reference answers.

\section{Experimental Setup} \label{sec:experimental_setup}

\newcommand{\modelNameStyle}[1]{\textsc{#1}}
\newcommand{\suffStyle}[1]{$_{\text{#1}}$}
\newcommand{\retrained}{Retr.}
\newcommand{\txl}{\modelNameStyle{TXL}}
\newcommand{\txlStale}{\modelNameStyle{TXL\suffStyle{Stale}}}
\newcommand{\txlIterative}{\modelNameStyle{TXL\suffStyle{FT}}}
\newcommand{\txlFT}{\txlIterative}
\newcommand{\txlRetrained}{\modelNameStyle{TXL\suffStyle{\retrained}}}

\newcommand{\cb}{\modelNameStyle{CB}}
\newcommand{\cbStale}{\modelNameStyle{CB\suffStyle{+Stale}}}
\newcommand{\cbFT}{\modelNameStyle{CB\suffStyle{+FT}}}
\newcommand{\cbRetrained}{\modelNameStyle{CB\suffStyle{+\retrained}}}
\newcommand{\ob}{\modelNameStyle{OB}}
\newcommand{\obIU}{\modelNameStyle{OB\suffStyle{+IU}}}
\newcommand{\obIUFT}{\modelNameStyle{OB\suffStyle{+IU+FT}}}
\newcommand{\obIURetrained}{\modelNameStyle{OB\suffStyle{+IU+\retrained}}}
\newcommand{\obGold}{\modelNameStyle{OB\suffStyle{+GoldRetr}}}
\newcommand{\obGoldFT}{\modelNameStyle{OB\suffStyle{+GoldRetr+FT}}}
\newcommand{\fid}{\modelNameStyle{FiD}}
\newcommand{\fidIU}{\modelNameStyle{FiD\suffStyle{+IU}}}
\newcommand{\fidIUFT}{\modelNameStyle{FiD\suffStyle{+IU+FT}}}
\newcommand{\fidIURetrained}{\modelNameStyle{FiD\suffStyle{+IU+\retrained}}}
\newcommand{\fidGold}{\modelNameStyle{FiD\suffStyle{+GoldRetr}}}
\newcommand{\fidGoldFT}{\modelNameStyle{FiD\suffStyle{+GoldRetr+FT}}}
\newcommand{\fidGoldRetrained}{\modelNameStyle{FiD\suffStyle{+GoldRetr+\retrained}}}

To evaluate how well parametric and semi-parametric approaches to QA ingest and understand unseen information, we consider
an auto-regressive, left-to-right Transformer-XL (\txl) \cite{dai-etal-2019-transformer} language model as our parametric, closed-book QA model (\cb), and use it as the underlying LM for our RAG-style \cite{NEURIPS2020_6b493230} semi-parametric, open-book QA model (\ob). We also consider a more recent open-book model based on Fusion-in-Decoder (\fid) that uses a T5 \cite{2020t5} sequence-to-sequence model\footnote{For T5, due to its existing pre-training, we control the knowledge of our model less precisely. However, both the T5 model and the BERT model (in DPR) were released in or before 2019, and so don't contain any knowledge from 2020, our evaluation period.}.
We use the standard metrics for QA: the F1 and the exact match (EM), after normalizing the answers, in the same way as \citet{rajpurkar-etal-2016-squad}. These are suitable for the on average short answers in our dataset.

To include the temporal metadata, we prefix publication dates to articles (``Thursday, February 7, 2019. [article text]''), and for questions we add question dates (``Today is Wednesday, May 6, 2020. [question text]'').

\subsection{Language Model Pretraining and Finetuning}
\label{sec:lms}

We consider three setups of \txl\ training:
\txlStale\ model is trained on WMT articles until the end of December 2019 (approx.\ 10.1M articles), and is missing knowledge required for answering questions in the recent subset.
\txlRetrained\ model is re-trained from scratch on all WMT articles until the end of December 2020, i.e., including the evaluation period (approx.\ 11.4M articles).
Lastly, \txlIterative\ model is \txlStale\ that we additionally \emph{iteratively fine-tune}\footnote{The best checkpoint of month $N$ is used as the starting point for fine-tuning on the articles of month $N+1$.}
on the 2020 monthly article sets (each approx.\ 100k articles). \txlStale\ and \txlRetrained\ are trained for 200k steps on 32 TPUv3, whereas fine-tuning is performed for 10k steps.

Our \txl\ has 18 layers and 1,280 hidden units, resulting in 448M parameters, roughly 30\% larger than GPT2-Medium and BERT-large.
We set the \txl\ sequence length to 1,024, and the memory cache length to 384 during model pre-training, and use a SentencePiece vocabulary of 50,259 tokens \cite{kudo-richardson-2018-sentencepiece}.

Analogously, for T5 base, %
we fine-tune the pre-trained vanilla T5 model \cite{2020t5} for 300k steps on WMT articles until the end of 2019, and subsequently iteratively fine-tune for 4k steps on the 2020 monthly splits. The retrained version is fine-tuned for 300k steps on articles until the end of 2020 starting from the vanilla T5 checkpoint.

\subsection{Closed-book QA}\label{sec:closed-bookqa-setup}

We use the closed-book QA task to examine a language model's knowledge. The task is to answer questions without any additional context provided, $p(a_i|q_i)$.
We fine-tune each of the pre-trained \txl\ LMs (stale, fine-tuned, and fully retrained) for question answering on the \streamingqa training set, using 4 TPUv3, and select the best checkpoint, as measured by F1, on the validation set; both sets contain knowledge and questions asked in or before 2019. The QA models, \cbStale, \cbFT, and \cbRetrained, are then subsequently evaluated on the evaluation sets from 2020. Answers are sampled using greedy decoding.

\begin{figure}
    \centering
    \includegraphics[width=0.49\linewidth]{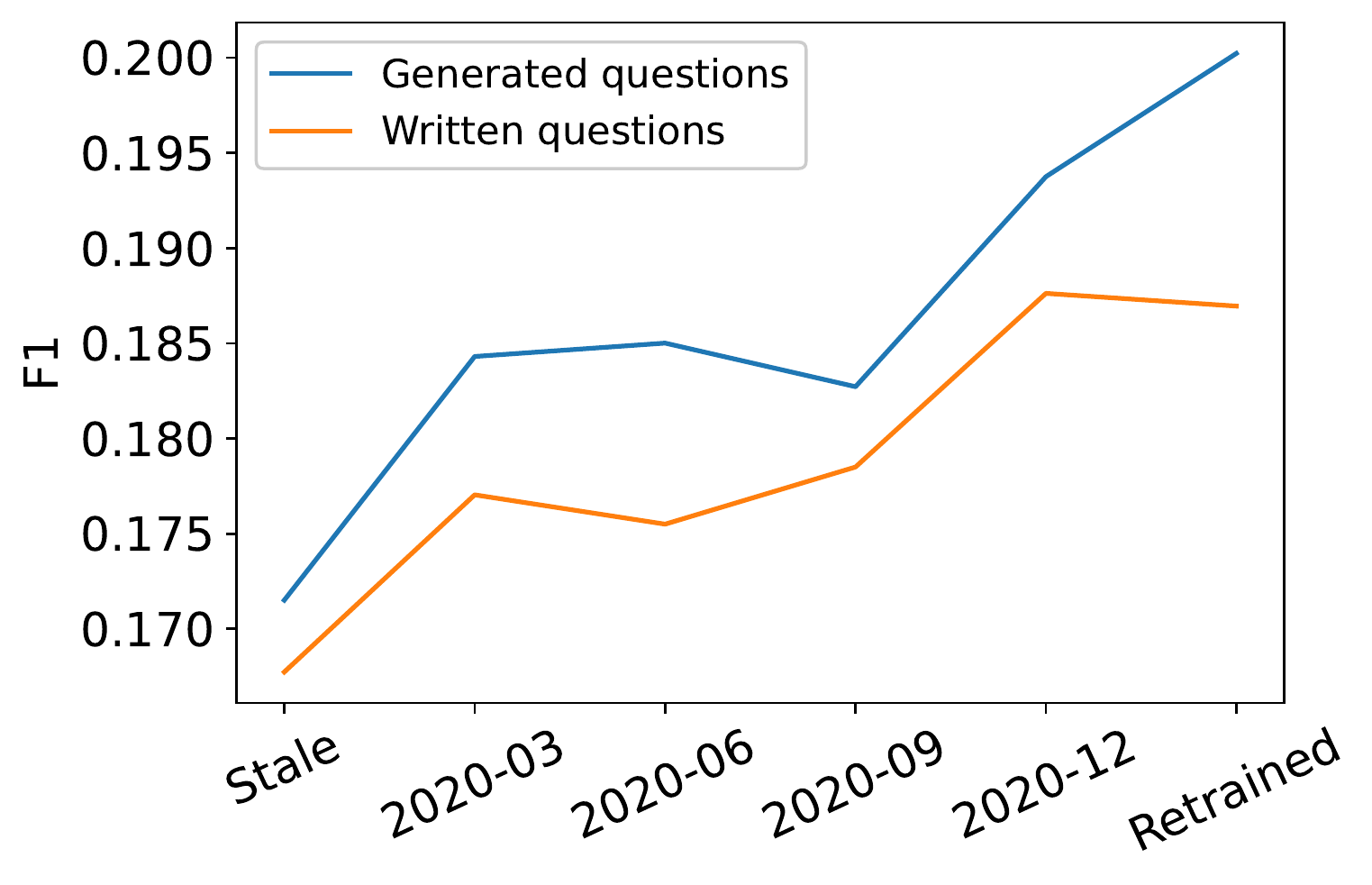}
    \includegraphics[width=0.49\linewidth]{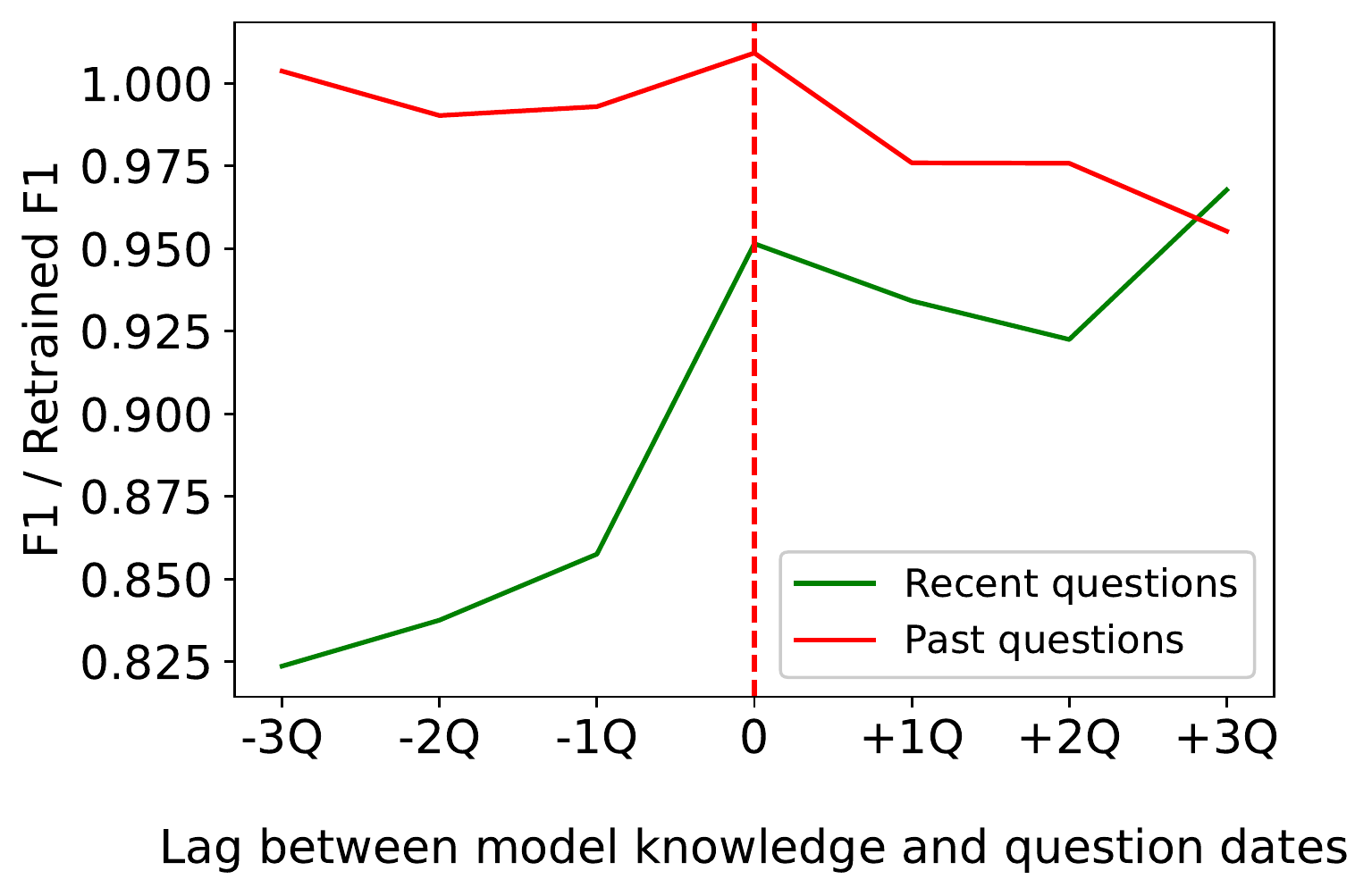}
    \caption{Left: F1 score on the whole evaluation dataset of \cbStale, \cbRetrained, and \cbFT\ fine-tuned on articles published until the specified cut-off dates.
    Right: The effect of a temporal lag between the final training month of \cbFT\ and question dates for generated questions, relative to \cbRetrained.}
    \label{fig:closed_book_overall}
    \label{fig:closed_book_temporal_lag}
\end{figure}

\subsection{Open-book QA}\label{sec:open-bookqa-setup}

In this task, we answer questions given a knowledge corpus of articles, $p(a_i|d_{q,i}, q_i, \mathcal C_{\leq t})$.
We use WMT news articles, sliced into 6-sentence chunks
as our knowledge corpus, resulting in
42.1M (up to 2019) and 47.6M (up to 2020) passages.

The \ob\ model is a variation of the Retrieval Augmented Generation model (RAG-sequence; \citet{NEURIPS2020_6b493230}) with a \txl-based\ generator, the same LMs as for our closed-book experiments.
As a retriever we use Dense Passage Retrieval (DPR; \citet{karpukhin-etal-2020-dense}), trained on question/passage pairs from our training set (including the question and publication dates), with embedding size of 768. We retrieve 20 passages.
We also consider the Fusion-in-Decoder model (\fid; \citet{izacard2020leveraging}), which was shown to outperform RAG on a number of QA tasks, with the same pre-trained DPR retriever as \ob\ but with 64 retrieved passages.
In contrast to RAG, the \fid's generator attends to all retrieved passages at the same time instead of individually.
We train both models with a restricted search space that contains gold evidence of all training and validation questions; this helps to reduce computation time (we did not see a material performance degradation due to this).

\section{Results and Analyses}

In this section we analyse the performance of the closed-book and open-book models on StreamingQA. We have three closed-book models: \cbStale, the iteratively finetuned \cbFT, and the retrained \cbRetrained. In order to adapt open-book models to new information from 2020, we always include new articles in the search index and then either keep the stale generator (\obIU, \fidIU), fine-tune the generator on new articles (\obIUFT, \fidIUFT), or use a retrained generator (\obIURetrained, \fidIURetrained).

\subsection{Adaptation to New Knowledge and Forgetting in Closed-Book QA}
\label{sec:results_cb}

\begin{figure*}
    \centering
    \includegraphics[width=0.49\linewidth]{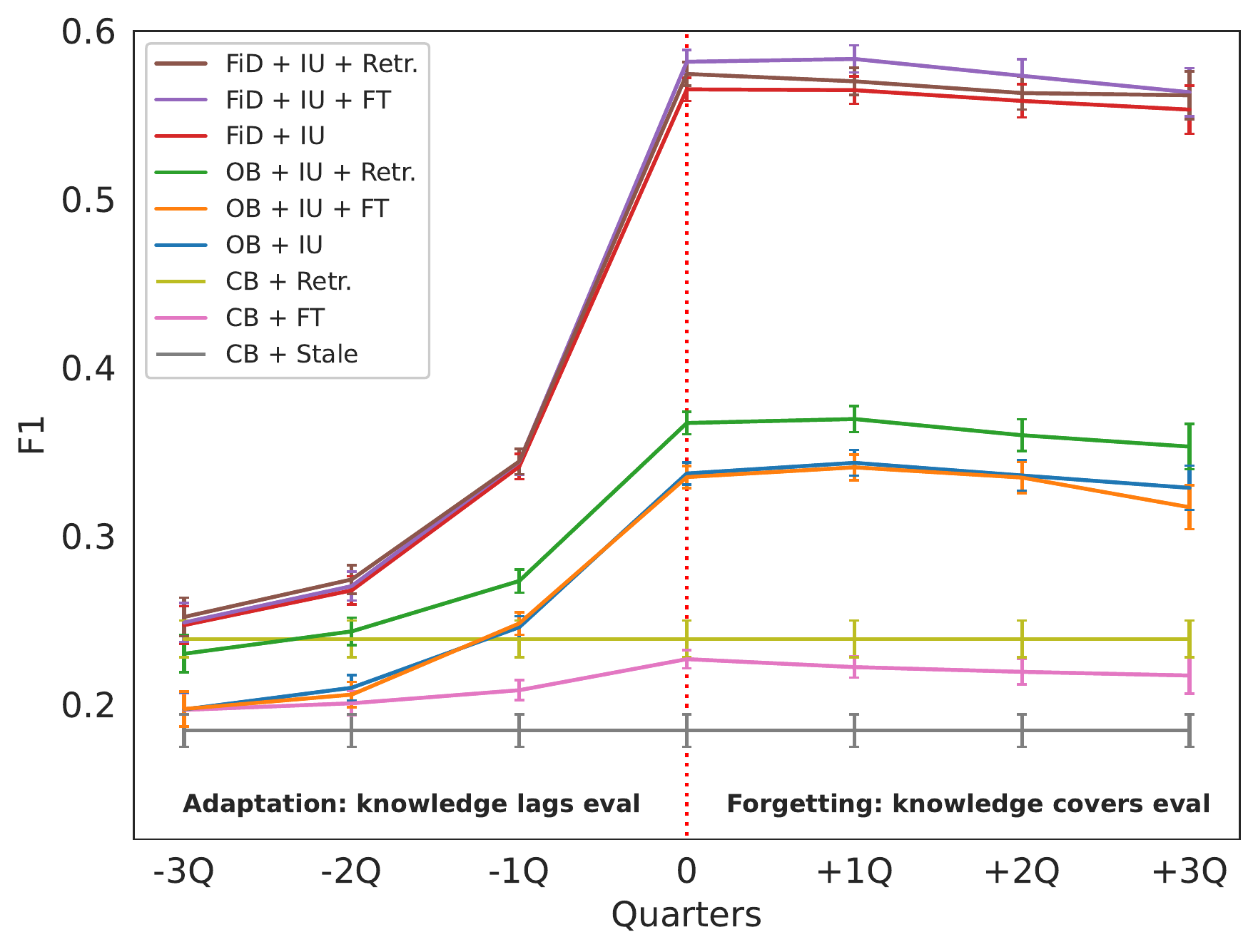}
    \includegraphics[width=0.49\linewidth]{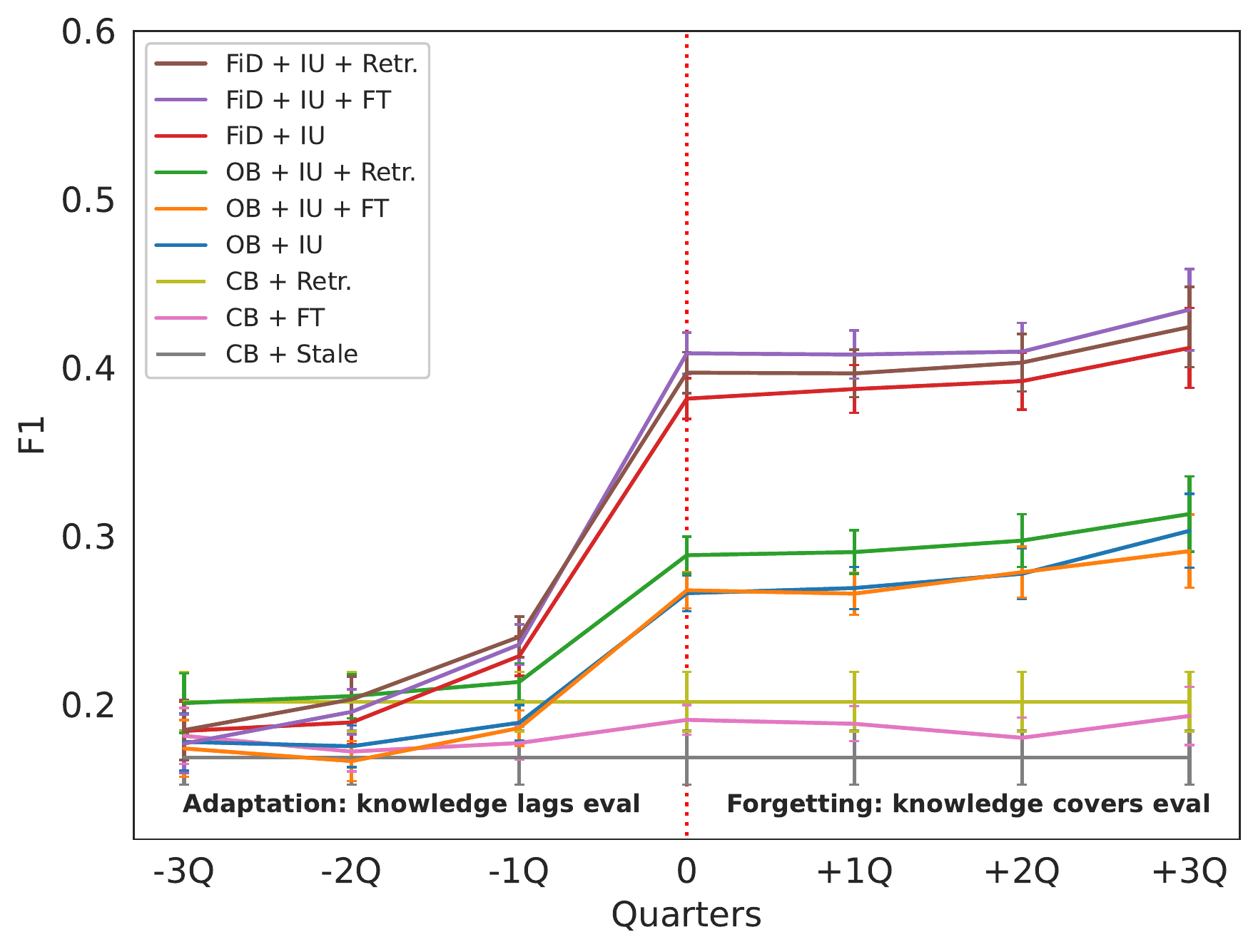}
    \caption{Adaptation and forgetting on recent subsets (generated, left; written, right).
    We observe that adapting the generator helps the FiD model, and helps the OB model when fully retrained, compared to index update only. Open-book models allow for much faster adaptation to recent knowledge than closed-book, with almost no forgetting. (IU = search index updated, FT = fine-tuned LM)}
    \label{fig:adaptation_recent}
\end{figure*}
\begin{figure}
    \centering
    \includegraphics[width=0.8\linewidth]{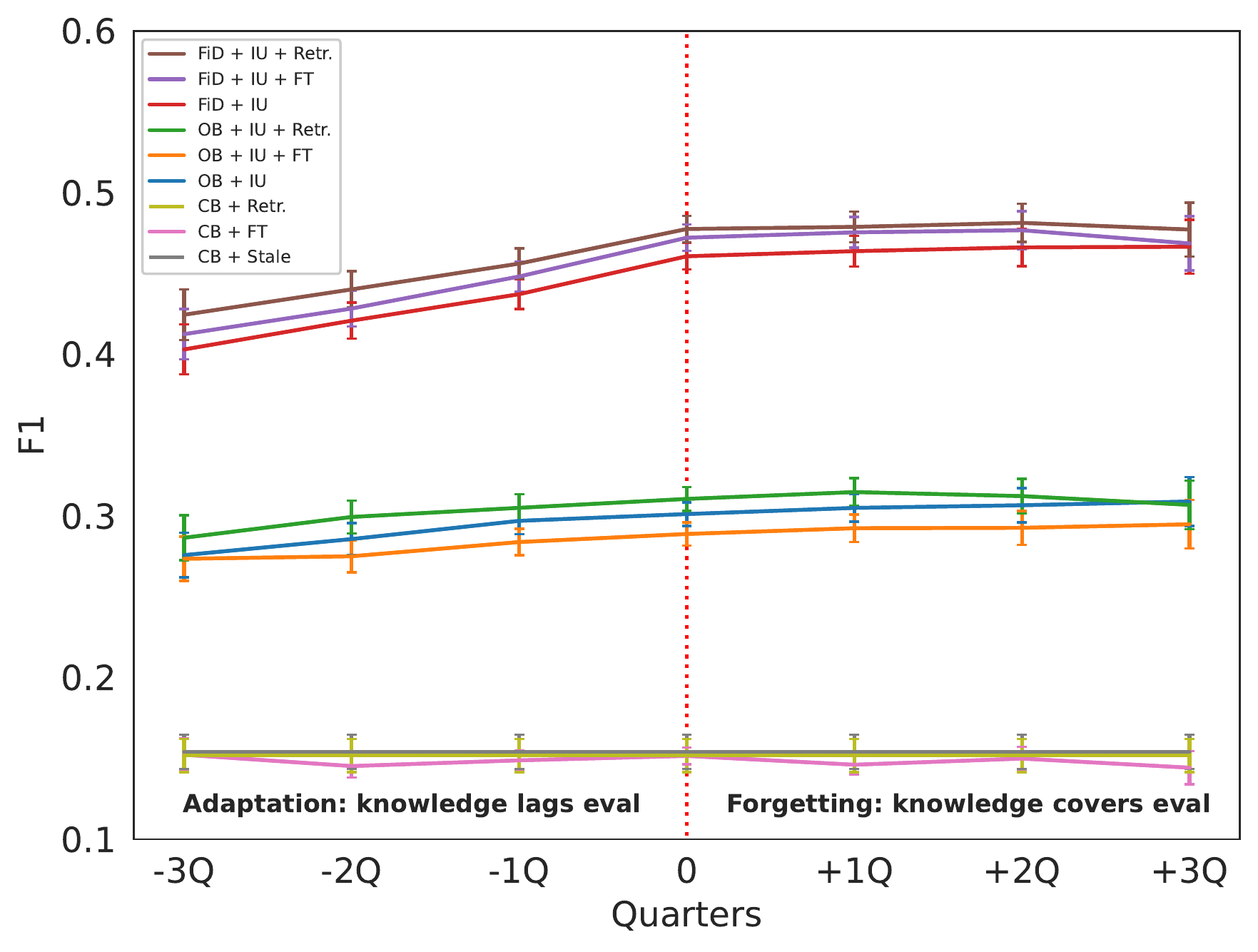}
    \caption{Adaptation and forgetting on past generated questions. We see only a slight improvement as the model acquires knowledge about 2020.
    We do not observe forgetting. %
    }
    \label{fig:adaptation_past_gen}
\end{figure}

\paragraph{Iterative LM fine-tuning improves performance on StreamingQA but lags retraining.}
We consider all questions in our evaluation sets (recent+past) and evaluate \cb\ models, in 
Figure~\ref{fig:closed_book_overall} (left). 
First, we observe that the \cbStale\ is outperformed by each of the \cbFT\ models---the fine-tuning is able to incorporate new information for the half of the questions that are only answerable from 2020 documents.
With each additional month of documents, the \cbFT\ models perform better for all answer types (named entities, phrases, dates; Appendix~\ref{sec:app_closedbook}). The improved performance is not simply due to
more data, but is driven by better accuracy on the recent subset, while the performance on the past subset remains mostly unchanged (Appendix~\ref{sec:app_closedbook}).
Secondly, we observe that \cbRetrained\ outperforms or is on par with all other models, and so vanilla adaptation that we consider for \cbFT\ should be improved
to bridge the gap from fine-tuning to retraining.

\textbf{Adaptation and forgetting}\hspace{0.5em}
We use question dates to split Eval-Generated and Eval-Written into quarterly sets. To understand how adaptation to new information is offset by forgetting of past information, we investigate the effect of a temporal lag between the question date and the end date of knowledge in the underlying LM.
Note that the question date and the knowledge date is on average much closer for the recent subset (a few weeks) compared to the past subset (years),
and so adaptation to new articles is more crucial for the recent subset performance.

When the lag is negative,
the model knowledge is lagging behind a question date (QD)---the model is missing necessary information\footnote{For example, an answer from a model trained until March 2020 for the question ``What does Donald Trump, US president, call his 2020 plan to expedite the development of a COVID-19 vaccine?'' asked on May 30, 2020 will be bucketed into -1Q.}. When the lag is positive, the questions are in the past with respect to the most recent information in the model, and in these settings, some previous information needed to answer these questions might have been overwritten---forgetting may occur.

We aggregate the model answers for each lag, and plot the corresponding F1 relative to that of \cbRetrained\ in Figure~\ref{fig:closed_book_temporal_lag} (right). 
As we fine-tune and the lag between the model knowledge and question month increases, the performance on the past subset slightly deteriorates until we under-perform \cbRetrained\ by about 5\%. 
On the recent subset, the performance first improves significantly,
and then as we pass the question quarter and continue fine-tuning on further data, we start seeing minor forgetting.
Similar conclusions hold for the written questions (Appendix~\ref{sec:app_closedbook}). %

\subsection{Adaptation to New Knowledge and Forgetting in Open-Book QA}
\label{sec:results_ob}

\textbf{Adaptation and forgetting}\hspace{0.5em} Similarly to the closed-book experiment,
we examine adaptation and forgetting by aggregating model answers by a temporal lag between the evaluation set and the end of the model knowledge. We consistently observe that on the recent subset of both generated and written questions (Figure~\ref{fig:adaptation_recent}), the open-book models (\ob, \fid) have a steep adaptation rate (from -1Q to 0Q) for all model variants, including just adding new articles into the search index without LM fine-tuning (\obIU, \fidIU).
For all of the models, we see almost no forgetting on the recent subsets.
In Figure~\ref{fig:adaptation_past_gen}, for generated past questions, there is no forgetting
(see Appendix~\ref{sec:app_openbook} for written questions).
Note that a small fraction of questions from the past subset reference articles from 2020, so seeing 2020 knowledge slightly helps compared to a lagging~model.

\textbf{Is updating search space sufficient, or do we need to update underlying LMs?}\hspace{0.5em}
Recent open-book QA models consist of a search index, a retriever, and a generator. One strong argument in favour of the open-book models is that new information can be added directly into the search index, potentially without any additional training.
Our results indicate that although the major performance gain comes from updating the search index, updating knowledge in the LM generator improves performance on recent subset questions too.
We compare \obIU, \fidIU, where the new documents are added only into the search index, with \obIUFT, \fidIUFT, \obIURetrained, \fidIURetrained, where both the index and the generator are updated.
We observe this in Figure~\ref{fig:adaptation_recent}, for lags of 0Q, 1Q, 2Q, and 3Q, where the models have the required knowledge for answering.
We see improvements for \fidIUFT\ (vs \fidIU) on the recent subset of generated and written questions, and on the past subset (Figure~\ref{fig:adaptation_past_gen}) at 0Q followed by minor forgetting. Moreover, retraining the generator improves performance for all models and all subsets.
For the \fid/T5 models, we see \fidIUFT\ performing somewhat better than \fidIURetrained\ on the recent generated questions and performing worse on the past subset, suggesting fine-tuning on the recent data improved performance on the corresponding knowledge.
Section~\ref{sec:results_param_vs_semiparam} explores why fine-tuning the generator is helping.

\begin{figure}
    \centering
    \includegraphics[width=\linewidth]{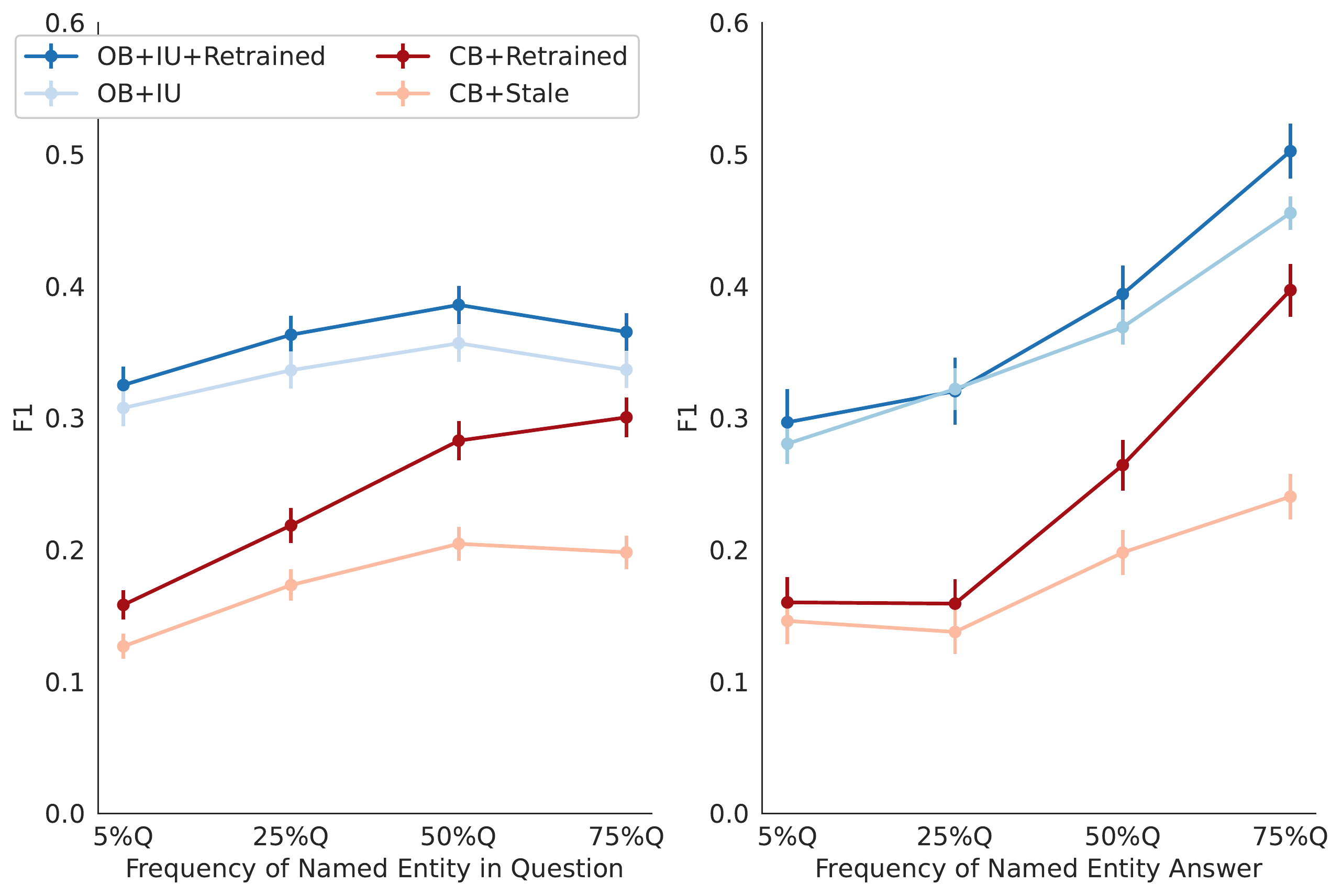}
    \vspace{-2mm}
    \caption{
    QA performance given question (left) or answer (right) named entity frequency quartiles.
    }
    \label{fig:frequency_analysis}
    \vspace{-1mm}
\end{figure}

\subsection{Parametric vs Semi-parametric Adaptation}
\label{sec:results_param_vs_semiparam}
Seeing above that fine-tuning or retraining the generator helps, we want to understand for which questions the updated generator is particularly important compared to a stale generator. \citet{lazaridou2021mind} previously demonstrated that one driving factor behind deteriorating temporal LM performance are changing frequencies of words, particularly named entities.
We analyze QA performance by %
frequency
of named entities appearing in (a) questions, and (b) answers, computed over the knowledge corpus up to 2019, and in 2020 only, respectively.

First, close-book performance is substantially better for questions that contain frequent named entities (see Appendix~\ref{sec:ap_dataset_examples} for examples): F1 is higher by absolute 10\% (Figure~\ref{fig:frequency_analysis}, left), likely due to higher-frequency named entities in the knowledge corpus providing a stronger learning signal for the parametric model.
Second, open-book performance does not show strong dependency on the frequency in the question, but this is due to two offsetting factors: DPR retrieval recall becomes worse with increasing frequency (recall@1 decreases by absolute 5\%), while the generator performance improves. Therefore, at lower frequencies better performance is driven by non-parametric adaptation through the updated search space, and at higher frequencies by parametric adaptation. 

Figure~\ref{fig:frequency_analysis} (right) shows performance as a function of answer named entity frequencies in 2020:
\cbRetrained\ outperforms \cbStale\ by 10-15\% for more frequent answers, suggesting that one reason for better performance of updated generators is more accurate modeling of word frequencies in 2020.

\subsection{Further Analyses}

\begin{figure}
    \centering
    \includegraphics[width=\linewidth]{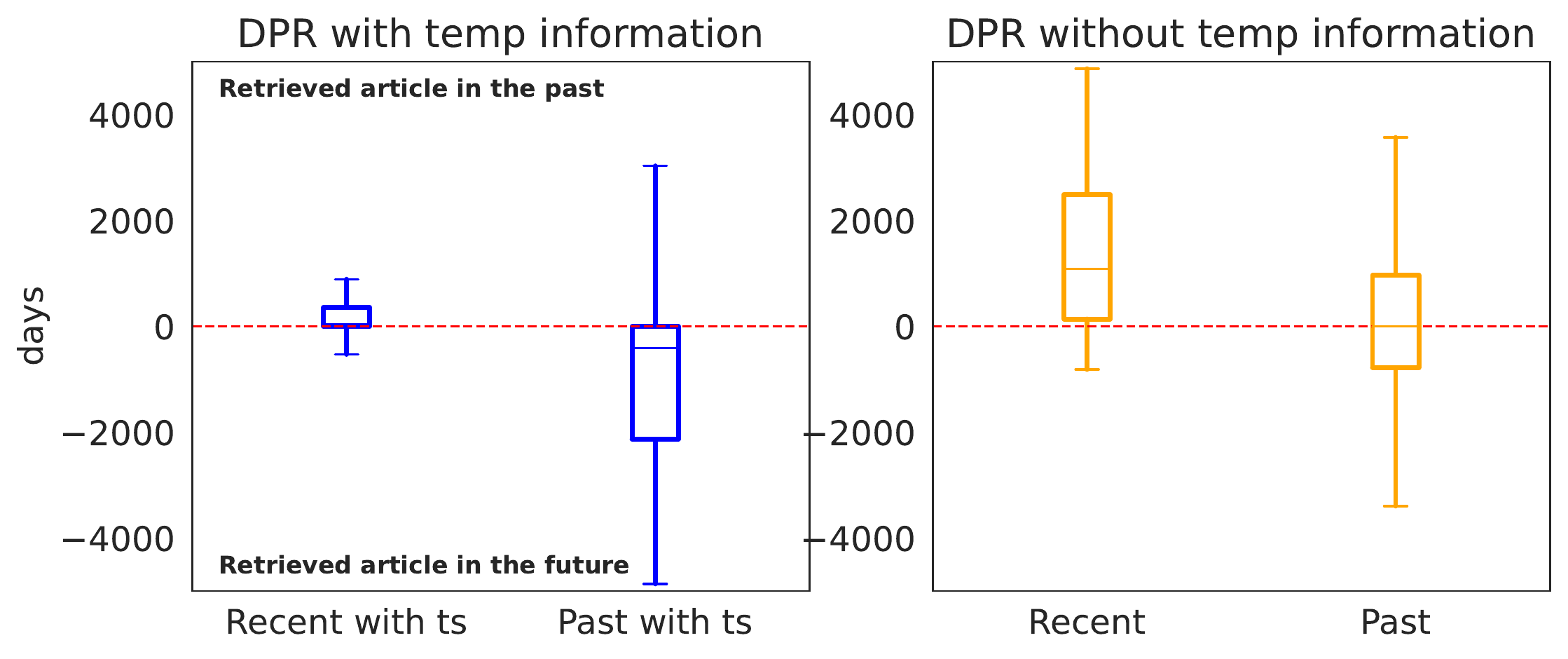}
    \vspace{-2mm}
    \caption{
    Temporal accuracy, measured in difference of days between gold passage timestamp (ts) minus retrieved passage ts, of the DPR trained on news for generated questions (recent vs past) with and without temporal information.
    }
    \label{fig:dpr_temporal_acc}
    \vspace{-2mm}
\end{figure}

\textbf{Temporal %
retrieval}\hspace{0.5em}
Using dates improves DPR performance for recent questions. In Figure~\ref{fig:dpr_temporal_acc}, median temporal difference between retrieved and gold articles is 41 and 1101 days for DPR with and without dates, respectively. Improved temporal accuracy translates into better recall overall, recall@20 for the recent generated questions is 57\% and 43\% for DPR with and without dates, respectively.
For past questions we do not see improvements, which suggests that the model may incorrectly interpret the two time specifications, i.e., the prepended question date and absolute or relative time specification in the question text.
See Appendix~\ref{sec:ap_tempretr} for more.

\textbf{Time specification in questions}\hspace{0.5em}
Generated questions in the past subset may contain an absolute or relative time specification in the question text, and we generally find that the open-book models perform best on questions without time specification\footnote{These question ask about the most recent past, a few weeks before the question dates.
}, followed by absolute, and relative. For example, for \fidIUFT, F1 is
0.711, 0.469, 0.359, respectively, and 0.441 overall.

\textbf{Static and updated questions}\hspace{0.5em}
For a preliminary analysis of ``static'' (knowledge that likely will not change) and ``updated'' (that might change) questions,
we observed that the open-book models generally performed worse on ``updated'' questions
(e.g., 0.435 vs 0.494 F1 for \fidIUFT\ on past, generated).
The evaluation sets have 6.7\%--11.2\% of likely static questions based on majority agreement. The recall of retrieved documents is better for the static questions.

\subsection{One-step Streaming and Static QA Benchmarks}
\label{sec:results_static}

\begin{figure}
    \centering
    \includegraphics[width=0.99\linewidth]{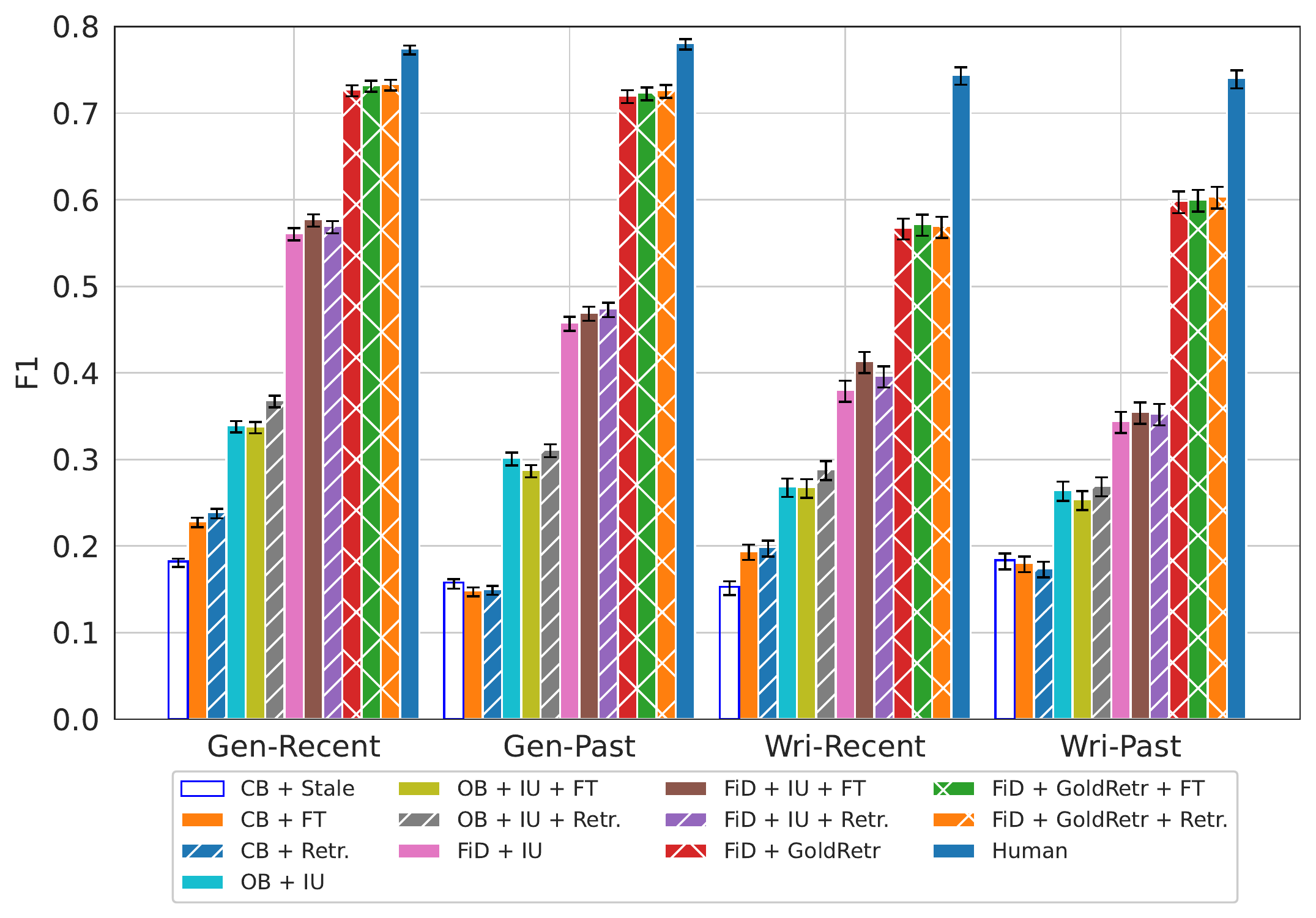}
    \vspace{-2mm}
    \caption{Static setup with all questions. Solid filled bars are models that had 2020 knowledge added incrementally.}
    \label{fig:static_all}
    \vspace{-2mm}
\end{figure}

\streamingqa dataset allows us to consider further two tasks,
and we provide benchmarks to encourage research on these directions:
one-step streaming setting\footnote{We report using the model fine-tuned iteratively on 12 months of news articles.} in Figure~\ref{fig:static_all} (solid bars), and the usual static open-book QA setup (diagonal-line bars),
evaluated on all 2020 questions.
There is still a large gap to human performance; moreover the dataset creates challenges to retrieval and news articles reading comprehension, see models with gold evidence versus retrieved (cross-pattern bars).
For the human benchmark we collected a fourth annotator answer.
See Appendix~\ref{sec:app_static} for EM and a table with all metrics.

\section{Related Work}

\textbf{QA datasets}\hspace{0.5em}
We summarize previous QA work on understanding of knowledge with temporal context in Section~\ref{sec:intro} and provide a dataset comparison table in Appendix~\ref{sec:app_datasets_table}.

\textbf{Question generation for QA}\hspace{0.2em}
Automatic question generation trained with supervision has been explored in QA for data augmentation \cite{alberti2019synthetic, dong2019unified,sultan-etal-2020-importance}, or as a way to enrich knowledge bases for QA-pair retriever models \cite{lewis2021paq}.
Here we instead leverage few-shot generation capabilities of large LMs
\cite{rae2021scaling,brown2020language}
to generate questions and use them for both training and evaluation. 

\textbf{Open-domain QA}\hspace{0.2em}
Progress in neural information retrieval \cite{karpukhin-etal-2020-dense, lee2019latent} enables open-domain QA models that are trained end-to-end as both the retriever and the reader are differentiable \cite{guu2020realm, NEURIPS2020_6b493230, izacard2020leveraging, sachan2021endtoend}. Recent work in the domain has focused on improving performance by combining information from multiple documents efficiently \cite{sachan2021endtoend, izacard2020leveraging} and on performance analysis of the dense retrievals, for instance, when dealing with named entities \cite{sciavolino2021simple, DBLP:journals/corr/abs-2109-01156}.

\textbf{Continual learning and distribution shift in LM and downstream tasks}\hspace{0.2em}
Continual learning in language is a long-standing research topic \cite{Carlson2010TowardAA, Parisi2018ContinualLL} that has recently seen an increase in interest. \citet{lazaridou2021mind} show that performance of Transformer-XL deteriorates when evaluated on data published after the training period, and use dynamic evaluation \cite{krause2017dynamic} to partially make up for this degradation. \citet{lazaridou2021mind} and \citet{hu2020drinking} release large scale benchmarks for studying temporal adaptation in the \emph{language modeling} task. \citet{jang2021continual} propose new metrics for knowledge updates and establish strong baselines. 
In contrast, we focus on studying adaptation in a downstream task of \emph{question answering}: we demonstrate that deterioration in perplexity translates into worse downstream performance and that adaptation through unsupervised fine-tuning or access to retrieval improves QA performance.  \citet{rottger2021temporal} study temporal adaptation of BERT models for the classification task and find that unsupervised temporal adaptation does not help downstream performance as much and task specific temporal adaptation is needed. \citet{hombaiah2021dynamic} propose new incremental methods for online BERT training using vocabulary expansion. In the context of semi-parametric models \citet{khandelwal2020generalization} and \citet{Lewis2020RetrievalAugmentedGF} describe flexible approaches to adaptation through updating information in the retrieval.

\section{Conclusion}

In order to enable a more realistic evaluation of QA models, we introduced the first QA dataset and task for studying adaptation to new information over time in open and close-book settings with temporally non-overlapping training and evaluation sets.

As language models grow bigger, the cost of maintaining them up-to-date increases, and therefore adaptation ability of the models becomes more important.
Our experimental results show that open-book QA models allow for fast and flexible adaptation through adding new articles into the search space,
with fine-tuning or retraining generally further improving performance. The ability to inject new knowledge through the search space depends on retrieval accuracy and the more up-to-date parametric LMs are capable of compensating for retrieval errors.
Additionally, our results show that iteratively fine-tuning the generator of the \fid\ QA model improves performance and costly retraining from scratch is not necessary.
We leave for future work to better understand and close the performance gap between retrained and stale generators.

\textbf{Future work}\hspace{0.3em}
\streamingqa\ highlights challenges of \emph{temporal reasoning} and invites further research into this area: the past subset contains questions with the relative time specification, where retrieval struggles to extract relevant passages.
For \emph{fine-tuning}, we consider a vanilla setup without delving deeply into more sophisticated continual learning approaches. Future work should take in-depths look at how best to adapt QA models, and the problem of what to compress into weights or what to add to the search space.
While we study adaptation to new knowledge, retrieving conflicting information due to \emph{updated knowledge} (e.g., ``How many seasons are in Game of Thrones?'') is another important direction we did not tackle here.

\section{Dataset Toxicity Discussion}

Toxic content is a concern in both human created and automatically generated content. We provide a discussion and describe our filtering of such content here. Our setup poses particular challenges as our questions and answers are based on news. First, answers in the dataset follow information in the articles regardless of the factual basis of the articles. While most of the news articles in WMT are from reputable news sources, news in general can contain content that may be considered toxic, such as graphic descriptions of crimes, or some quotes or opinions. Second, as some questions and answers are generated using a large language model, there is a risk that it may generate toxic content; however, we want to note that the generation process is constrained by conditioning on the article and a substring answer and the subsequent automatic filtering. Third, our dataset overall is intended to evaluate adaptation of models to new information in news over time, and therefore, it may not be applicable to settings  where the assumptions we made don't apply.

We aimed to create a balanced process that identifies most of the toxic content while decreasing the risk of removing false positives.
To identify toxic content, we used the Perspective API\footnote{\url{https://perspectiveapi.com/}} which provides classifiers for several categories of toxic content (identity attack, insult, threat, profanity, sexually explicit, severe toxicity). We decided to use the specific classifiers instead of the generic toxicity classifier because our initial annotations indicated that the specific classifiers perform better. Removing content needs to be done with care as these classifiers do contain false positives (e.g., people, [Republic of] Niger, shoot, death, abuse, balls [in sports], and [last] names which bear phonetic similarity to insults), and removing too many such examples may cause harm by decreasing representation of some groups (e.g., black, muslim, jewish, LGBTQIA+ minorities). Through manual annotation of the questions with the highest toxicity scores, we have determined thresholds for removing questions as follows: for each 0.05 band of the scores from 1 to 0 (e.g., [1.0, 0.95], [0.95, 0.90], …), we remove questions in each band until two subsequent bands contain fewer than 30\% toxic questions (judged by two annotators on a sample of 50 per band). The first of these two subsequent bands is also removed. We annotated more than 5.5k examples throughout this process.
As the annotation judgements for filtering were done by a small group of annotators, we cannot claim that the annotation had perfect representivity nor that the annotators had full cultural context from all possible views.
For our manual annotation, we adapted the Perspective API classifier definitions in a minor way (see Appendix~\ref{ap:toxicity}). This filtering resulted in removing about 0.57\% of questions (0.60\%, 0.61\%, 0.43\%, 0.65\% from Train, Valid, Eval-Generated, Eval-Written), and thresholds of 0.75 for identity attack, 0.80 for insult, 0.65 for profanity, 0.55 for severe toxicity, 0.85 for sexually explicit, and 0.90 for threat. Subsequently, we estimated that 0.5\% of toxic questions remain (sample of 1k questions). We provide the automatic toxicity scores as part of the data release. This approach was formed with input from DeepMind's ethics and safety teams, and with guidance from our multidisciplinary leadership group which advises on societal impacts associated with research.

\section*{Acknowledgements}

We thank our human annotators for helping create a large part of the dataset.
We also thank John Aslanides and Kevin McKee for advice on the initial setup of the human annotation data collection.
We would particularly like to thank Boxi Wu for her support in helping to organize panels with experts to advise on our toxicity filtering approach.
Lastly, we would like to acknowledge Dani Yogatama and Lisa Anne Hendricks who acted as our internal reviewers and thank Chris Dyer for his continuous input.

\bibliography{references}

\begin{thebibliography}{40}
\providecommand{\natexlab}[1]{#1}
\providecommand{\url}[1]{\texttt{#1}}
\expandafter\ifx\csname urlstyle\endcsname\relax
  \providecommand{\doi}[1]{doi: #1}\else
  \providecommand{\doi}{doi: \begingroup \urlstyle{rm}\Url}\fi

\bibitem[Akhbardeh et~al.(2021)Akhbardeh, Arkhangorodsky, Biesialska, Bojar,
  Chatterjee, Chaudhary, Costa-jussà, España-Bonet, Fan, Federman, Freitag,
  Graham, Grundkiewicz, Haddow, Harter, Heafield, Homan, Huck,
  Amponsah-Kaakyire, Kasai, Khashabi, Knight, Kocmi, Koehn, Lourie, Monz,
  Morishita, Nagata, Nagesh, Nakazawa, Negri, Pal, Tapo, Turchi, Vydrin, and
  Zampieri]{akhbardeh2021wmt}
Akhbardeh, F., Arkhangorodsky, A., Biesialska, M., Bojar, O., Chatterjee, R.,
  Chaudhary, V., Costa-jussà, M.~R., España-Bonet, C., Fan, A., Federman, C.,
  Freitag, M., Graham, Y., Grundkiewicz, R., Haddow, B., Harter, L., Heafield,
  K., Homan, C.~M., Huck, M., Amponsah-Kaakyire, K., Kasai, J., Khashabi, D.,
  Knight, K., Kocmi, T., Koehn, P., Lourie, N., Monz, C., Morishita, M.,
  Nagata, M., Nagesh, A., Nakazawa, T., Negri, M., Pal, S., Tapo, A., Turchi,
  M., Vydrin, V., and Zampieri, M.
\newblock Findings of the 2021 conference on machine translation (wmt21).
\newblock In \emph{Proceedings of the Sixth Conference on Machine Translation},
  pp.\  1--88, Online, 2021.

\bibitem[Alberti et~al.(2019)Alberti, Andor, Pitler, Devlin, and
  Collins]{alberti2019synthetic}
Alberti, C., Andor, D., Pitler, E., Devlin, J., and Collins, M.
\newblock Synthetic qa corpora generation with roundtrip consistency, 2019.

\bibitem[Amba~Hombaiah et~al.(2021)Amba~Hombaiah, Chen, Zhang, Bendersky, and
  Najork]{hombaiah2021dynamic}
Amba~Hombaiah, S., Chen, T., Zhang, M., Bendersky, M., and Najork, M.
\newblock Dynamic language models for continuously evolving content.
\newblock \emph{Proceedings of the 27th ACM SIGKDD Conference on Knowledge
  Discovery \& Data Mining}, Aug 2021.
\newblock \doi{10.1145/3447548.3467162}.
\newblock URL \url{http://dx.doi.org/10.1145/3447548.3467162}.

\bibitem[Brown et~al.(2020)Brown, Mann, Ryder, Subbiah, Kaplan, Dhariwal,
  Neelakantan, Shyam, Sastry, Askell, Agarwal, Herbert-Voss, Krueger, Henighan,
  Child, Ramesh, Ziegler, Wu, Winter, Hesse, Chen, Sigler, Litwin, Gray, Chess,
  Clark, Berner, McCandlish, Radford, Sutskever, and Amodei]{brown2020language}
Brown, T.~B., Mann, B., Ryder, N., Subbiah, M., Kaplan, J., Dhariwal, P.,
  Neelakantan, A., Shyam, P., Sastry, G., Askell, A., Agarwal, S.,
  Herbert-Voss, A., Krueger, G., Henighan, T., Child, R., Ramesh, A., Ziegler,
  D.~M., Wu, J., Winter, C., Hesse, C., Chen, M., Sigler, E., Litwin, M., Gray,
  S., Chess, B., Clark, J., Berner, C., McCandlish, S., Radford, A., Sutskever,
  I., and Amodei, D.
\newblock Language models are few-shot learners, 2020.

\bibitem[Carlson et~al.(2010)Carlson, Betteridge, Kisiel, Settles, Hruschka,
  and Mitchell]{Carlson2010TowardAA}
Carlson, A., Betteridge, J., Kisiel, B., Settles, B., Hruschka, E.~R., and
  Mitchell, T.~M.
\newblock Toward an architecture for never-ending language learning.
\newblock In \emph{AAAI}, 2010.

\bibitem[Chen et~al.(2021)Chen, Wang, and Wang]{chen2021dataset}
Chen, W., Wang, X., and Wang, W.~Y.
\newblock A dataset for answering time-sensitive questions, 2021.

\bibitem[Dai et~al.(2019)Dai, Yang, Yang, Carbonell, Le, and
  Salakhutdinov]{dai-etal-2019-transformer}
Dai, Z., Yang, Z., Yang, Y., Carbonell, J., Le, Q., and Salakhutdinov, R.
\newblock Transformer-{XL}: Attentive language models beyond a fixed-length
  context.
\newblock In \emph{Proceedings of the 57th Annual Meeting of the Association
  for Computational Linguistics}, pp.\  2978--2988, Florence, Italy, July 2019.
  Association for Computational Linguistics.
\newblock \doi{10.18653/v1/P19-1285}.
\newblock URL \url{https://aclanthology.org/P19-1285}.

\bibitem[Dhingra et~al.(2021)Dhingra, Cole, Eisenschlos, Gillick, Eisenstein,
  and Cohen]{dhingra2021timeaware}
Dhingra, B., Cole, J.~R., Eisenschlos, J.~M., Gillick, D., Eisenstein, J., and
  Cohen, W.~W.
\newblock Time-aware language models as temporal knowledge bases, 2021.

\bibitem[Dong et~al.(2019)Dong, Yang, Wang, Wei, Liu, Wang, Gao, Zhou, and
  Hon]{dong2019unified}
Dong, L., Yang, N., Wang, W., Wei, F., Liu, X., Wang, Y., Gao, J., Zhou, M.,
  and Hon, H.-W.
\newblock Unified language model pre-training for natural language
  understanding and generation, 2019.

\bibitem[Guu et~al.(2020)Guu, Lee, Tung, Pasupat, and Chang]{guu2020realm}
Guu, K., Lee, K., Tung, Z., Pasupat, P., and Chang, M.-W.
\newblock Realm: Retrieval-augmented language model pre-training, 2020.

\bibitem[Hu et~al.(2020)Hu, Sener, Sha, and Koltun]{hu2020drinking}
Hu, H., Sener, O., Sha, F., and Koltun, V.
\newblock Drinking from a firehose: Continual learning with web-scale natural
  language, 2020.

\bibitem[Izacard \& Grave(2020)Izacard and Grave]{izacard2020leveraging}
Izacard, G. and Grave, E.
\newblock Leveraging passage retrieval with generative models for open domain
  question answering, 2020.

\bibitem[Jang et~al.(2021)Jang, Ye, Yang, Shin, Han, Kim, Choi, and
  Seo]{jang2021continual}
Jang, J., Ye, S., Yang, S., Shin, J., Han, J., Kim, G., Choi, S.~J., and Seo,
  M.
\newblock Towards continual knowledge learning of language models, 2021.

\bibitem[Jia et~al.(2018)Jia, Abujabal, Saha~Roy, Str\"{o}tgen, and
  Weikum]{10.1145/3184558.3191536}
Jia, Z., Abujabal, A., Saha~Roy, R., Str\"{o}tgen, J., and Weikum, G.
\newblock Tempquestions: A benchmark for temporal question answering.
\newblock In \emph{Companion Proceedings of the The Web Conference 2018}, WWW
  '18, pp.\  1057–1062, Republic and Canton of Geneva, CHE, 2018.
  International World Wide Web Conferences Steering Committee.
\newblock ISBN 9781450356404.
\newblock \doi{10.1145/3184558.3191536}.
\newblock URL \url{https://doi.org/10.1145/3184558.3191536}.

\bibitem[Jia et~al.(2021)Jia, Pramanik, Saha~Roy, and
  Weikum]{timequestions2021}
Jia, Z., Pramanik, S., Saha~Roy, R., and Weikum, G.
\newblock Complex temporal question answering on knowledge graphs.
\newblock \emph{Proceedings of the 30th ACM International Conference on
  Information \& Knowledge Management}, Oct 2021.
\newblock \doi{10.1145/3459637.3482416}.
\newblock URL \url{http://dx.doi.org/10.1145/3459637.3482416}.

\bibitem[Karpukhin et~al.(2020)Karpukhin, Oguz, Min, Lewis, Wu, Edunov, Chen,
  and Yih]{karpukhin-etal-2020-dense}
Karpukhin, V., Oguz, B., Min, S., Lewis, P., Wu, L., Edunov, S., Chen, D., and
  Yih, W.-t.
\newblock Dense passage retrieval for open-domain question answering.
\newblock In \emph{Proceedings of the 2020 Conference on Empirical Methods in
  Natural Language Processing (EMNLP)}, pp.\  6769--6781, Online, November
  2020. Association for Computational Linguistics.
\newblock \doi{10.18653/v1/2020.emnlp-main.550}.
\newblock URL \url{https://aclanthology.org/2020.emnlp-main.550}.

\bibitem[Khandelwal et~al.(2020)Khandelwal, Levy, Jurafsky, Zettlemoyer, and
  Lewis]{khandelwal2020generalization}
Khandelwal, U., Levy, O., Jurafsky, D., Zettlemoyer, L., and Lewis, M.
\newblock Generalization through memorization: Nearest neighbor language
  models.
\newblock In \emph{International Conference on Learning Representations}, 2020.
\newblock URL \url{https://openreview.net/forum?id=HklBjCEKvH}.

\bibitem[Ko{\v{c}}isk{\'y} et~al.(2018)Ko{\v{c}}isk{\'y}, Schwarz, Blunsom,
  Dyer, Hermann, Melis, and Grefenstette]{kocisky-etal-2018-narrativeqa}
Ko{\v{c}}isk{\'y}, T., Schwarz, J., Blunsom, P., Dyer, C., Hermann, K.~M.,
  Melis, G., and Grefenstette, E.
\newblock The {N}arrative{QA} reading comprehension challenge.
\newblock \emph{Transactions of the Association for Computational Linguistics},
  6:\penalty0 317--328, 2018.
\newblock \doi{10.1162/tacl_a_00023}.
\newblock URL \url{https://aclanthology.org/Q18-1023}.

\bibitem[Krause et~al.(2017)Krause, Kahembwe, Murray, and
  Renals]{krause2017dynamic}
Krause, B., Kahembwe, E., Murray, I., and Renals, S.
\newblock Dynamic evaluation of neural sequence models, 2017.

\bibitem[Kudo \& Richardson(2018)Kudo and
  Richardson]{kudo-richardson-2018-sentencepiece}
Kudo, T. and Richardson, J.
\newblock {S}entence{P}iece: A simple and language independent subword
  tokenizer and detokenizer for neural text processing.
\newblock In \emph{Proceedings of the 2018 Conference on Empirical Methods in
  Natural Language Processing: System Demonstrations}, pp.\  66--71, Brussels,
  Belgium, November 2018. Association for Computational Linguistics.
\newblock \doi{10.18653/v1/D18-2012}.
\newblock URL \url{https://aclanthology.org/D18-2012}.

\bibitem[Kwiatkowski et~al.(2019)Kwiatkowski, Palomaki, Redfield, Collins,
  Parikh, Alberti, Epstein, Polosukhin, Kelcey, Devlin, Lee, Toutanova, Jones,
  Chang, Dai, Uszkoreit, Le, and Petrov]{47761-NaturalQuestions}
Kwiatkowski, T., Palomaki, J., Redfield, O., Collins, M., Parikh, A., Alberti,
  C., Epstein, D., Polosukhin, I., Kelcey, M., Devlin, J., Lee, K., Toutanova,
  K.~N., Jones, L., Chang, M.-W., Dai, A., Uszkoreit, J., Le, Q., and Petrov,
  S.
\newblock Natural questions: a benchmark for question answering research.
\newblock \emph{Transactions of the Association of Computational Linguistics},
  2019.

\bibitem[Lazaridou et~al.(2021)Lazaridou, Kuncoro, Gribovskaya, Agrawal,
  Li{\v{s}}ka, Terzi, Gimenez, de~Masson~d'Autume, Ko{\v{c}}isk{\'y}, Ruder,
  Yogatama, Cao, Young, and Blunsom]{lazaridou2021mind}
Lazaridou, A., Kuncoro, A., Gribovskaya, E., Agrawal, D., Li{\v{s}}ka, A.,
  Terzi, T., Gimenez, M., de~Masson~d'Autume, C., Ko{\v{c}}isk{\'y}, T., Ruder,
  S., Yogatama, D., Cao, K., Young, S., and Blunsom, P.
\newblock Mind the gap: Assessing temporal generalization in neural language
  models.
\newblock In \emph{Thirty-Fifth Conference on Neural Information Processing
  Systems}, 2021.
\newblock URL \url{https://openreview.net/forum?id=73OmmrCfSyy}.

\bibitem[Lee et~al.(2019{\natexlab{a}})Lee, Chang, and
  Toutanova]{lee-etal-2019-latent-NQ-OPEN}
Lee, K., Chang, M.-W., and Toutanova, K.
\newblock Latent retrieval for weakly supervised open domain question
  answering.
\newblock In \emph{Proceedings of the 57th Annual Meeting of the Association
  for Computational Linguistics}, pp.\  6086--6096, Florence, Italy, July
  2019{\natexlab{a}}. Association for Computational Linguistics.
\newblock \doi{10.18653/v1/P19-1612}.
\newblock URL \url{https://aclanthology.org/P19-1612}.

\bibitem[Lee et~al.(2019{\natexlab{b}})Lee, Chang, and
  Toutanova]{lee2019latent}
Lee, K., Chang, M.-W., and Toutanova, K.
\newblock Latent retrieval for weakly supervised open domain question
  answering, 2019{\natexlab{b}}.

\bibitem[Lewis et~al.(2020{\natexlab{a}})Lewis, Perez, Piktus, Petroni,
  Karpukhin, Goyal, Kuttler, Lewis, tau Yih, Rockt{\"a}schel, Riedel, and
  Kiela]{Lewis2020RetrievalAugmentedGF}
Lewis, P., Perez, E., Piktus, A., Petroni, F., Karpukhin, V., Goyal, N.,
  Kuttler, H., Lewis, M., tau Yih, W., Rockt{\"a}schel, T., Riedel, S., and
  Kiela, D.
\newblock Retrieval-augmented generation for knowledge-intensive nlp tasks.
\newblock \emph{ArXiv}, abs/2005.11401, 2020{\natexlab{a}}.

\bibitem[Lewis et~al.(2020{\natexlab{b}})Lewis, Perez, Piktus, Petroni,
  Karpukhin, Goyal, K\"{u}ttler, Lewis, Yih, Rockt\"{a}schel, Riedel, and
  Kiela]{NEURIPS2020_6b493230}
Lewis, P., Perez, E., Piktus, A., Petroni, F., Karpukhin, V., Goyal, N.,
  K\"{u}ttler, H., Lewis, M., Yih, W.-t., Rockt\"{a}schel, T., Riedel, S., and
  Kiela, D.
\newblock Retrieval-augmented generation for knowledge-intensive nlp tasks.
\newblock In Larochelle, H., Ranzato, M., Hadsell, R., Balcan, M.~F., and Lin,
  H. (eds.), \emph{Advances in Neural Information Processing Systems},
  volume~33, pp.\  9459--9474. Curran Associates, Inc., 2020{\natexlab{b}}.
\newblock URL
  \url{https://proceedings.neurips.cc/paper/2020/file/6b493230205f780e1bc26945df7481e5-Paper.pdf}.

\bibitem[Lewis et~al.(2021)Lewis, Wu, Liu, Minervini, Küttler, Piktus,
  Stenetorp, and Riedel]{lewis2021paq}
Lewis, P., Wu, Y., Liu, L., Minervini, P., Küttler, H., Piktus, A., Stenetorp,
  P., and Riedel, S.
\newblock Paq: 65 million probably-asked questions and what you can do with
  them, 2021.

\bibitem[Liu et~al.(2021)Liu, Lewis, Riedel, and
  Stenetorp]{DBLP:journals/corr/abs-2109-01156}
Liu, L., Lewis, P. S.~H., Riedel, S., and Stenetorp, P.
\newblock Challenges in generalization in open domain question answering.
\newblock \emph{CoRR}, abs/2109.01156, 2021.
\newblock URL \url{https://arxiv.org/abs/2109.01156}.

\bibitem[Ning et~al.(2020)Ning, Wu, Han, Peng, Gardner, and
  Roth]{ning-etal-2020-torque}
Ning, Q., Wu, H., Han, R., Peng, N., Gardner, M., and Roth, D.
\newblock {TORQUE}: A reading comprehension dataset of temporal ordering
  questions.
\newblock In \emph{Proceedings of the 2020 Conference on Empirical Methods in
  Natural Language Processing (EMNLP)}, pp.\  1158--1172, Online, November
  2020. Association for Computational Linguistics.
\newblock \doi{10.18653/v1/2020.emnlp-main.88}.
\newblock URL \url{https://aclanthology.org/2020.emnlp-main.88}.

\bibitem[Parisi et~al.(2018)Parisi, Kemker, Part, Kanan, and
  Wermter]{Parisi2018ContinualLL}
Parisi, G.~I., Kemker, R., Part, J.~L., Kanan, C., and Wermter, S.
\newblock Continual lifelong learning with neural networks: A review.
\newblock 2018.

\bibitem[Rae et~al.(2021)Rae, Borgeaud, Cai, Millican, Hoffmann, Song,
  Aslanides, Henderson, Ring, Young, et~al.]{rae2021scaling}
Rae, J.~W., Borgeaud, S., Cai, T., Millican, K., Hoffmann, J., Song, F.,
  Aslanides, J., Henderson, S., Ring, R., Young, S., et~al.
\newblock Scaling language models: Methods, analysis \& insights from training
  gopher.
\newblock \emph{arXiv preprint arXiv:2112.11446}, 2021.

\bibitem[Raffel et~al.(2020)Raffel, Shazeer, Roberts, Lee, Narang, Matena,
  Zhou, Li, and Liu]{2020t5}
Raffel, C., Shazeer, N., Roberts, A., Lee, K., Narang, S., Matena, M., Zhou,
  Y., Li, W., and Liu, P.~J.
\newblock Exploring the limits of transfer learning with a unified text-to-text
  transformer.
\newblock \emph{Journal of Machine Learning Research}, 21\penalty0
  (140):\penalty0 1--67, 2020.
\newblock URL \url{http://jmlr.org/papers/v21/20-074.html}.

\bibitem[Rajpurkar et~al.(2016)Rajpurkar, Zhang, Lopyrev, and
  Liang]{rajpurkar-etal-2016-squad}
Rajpurkar, P., Zhang, J., Lopyrev, K., and Liang, P.
\newblock {SQ}u{AD}: 100,000+ questions for machine comprehension of text.
\newblock In \emph{Proceedings of the 2016 Conference on Empirical Methods in
  Natural Language Processing}, pp.\  2383--2392, Austin, Texas, November 2016.
  Association for Computational Linguistics.
\newblock \doi{10.18653/v1/D16-1264}.
\newblock URL \url{https://aclanthology.org/D16-1264}.

\bibitem[R{\"o}ttger \& Pierrehumbert(2021)R{\"o}ttger and
  Pierrehumbert]{rottger2021temporal}
R{\"o}ttger, P. and Pierrehumbert, J.
\newblock Temporal adaptation of {BERT} and performance on downstream document
  classification: Insights from social media.
\newblock In \emph{Findings of the Association for Computational Linguistics:
  EMNLP 2021}, pp.\  2400--2412, Punta Cana, Dominican Republic, November 2021.
  Association for Computational Linguistics.
\newblock \doi{10.18653/v1/2021.findings-emnlp.206}.
\newblock URL \url{https://aclanthology.org/2021.findings-emnlp.206}.

\bibitem[Sachan et~al.(2021)Sachan, Reddy, Hamilton, Dyer, and
  Yogatama]{sachan2021endtoend}
Sachan, D.~S., Reddy, S., Hamilton, W., Dyer, C., and Yogatama, D.
\newblock End-to-end training of multi-document reader and retriever for
  open-domain question answering, 2021.

\bibitem[Saxena et~al.(2021)Saxena, Chakrabarti, and
  Talukdar]{saxena-etal-2021-question}
Saxena, A., Chakrabarti, S., and Talukdar, P.
\newblock Question answering over temporal knowledge graphs.
\newblock In \emph{Proceedings of the 59th Annual Meeting of the Association
  for Computational Linguistics and the 11th International Joint Conference on
  Natural Language Processing (Volume 1: Long Papers)}, pp.\  6663--6676,
  Online, August 2021. Association for Computational Linguistics.
\newblock \doi{10.18653/v1/2021.acl-long.520}.
\newblock URL \url{https://aclanthology.org/2021.acl-long.520}.

\bibitem[Sciavolino et~al.(2021)Sciavolino, Zhong, Lee, and
  Chen]{sciavolino2021simple}
Sciavolino, C., Zhong, Z., Lee, J., and Chen, D.
\newblock Simple entity-centric questions challenge dense retrievers, 2021.

\bibitem[Sultan et~al.(2020)Sultan, Chandel, Fernandez~Astudillo, and
  Castelli]{sultan-etal-2020-importance}
Sultan, M.~A., Chandel, S., Fernandez~Astudillo, R., and Castelli, V.
\newblock On the importance of diversity in question generation for {QA}.
\newblock In \emph{Proceedings of the 58th Annual Meeting of the Association
  for Computational Linguistics}, pp.\  5651--5656, Online, July 2020.
  Association for Computational Linguistics.
\newblock \doi{10.18653/v1/2020.acl-main.500}.
\newblock URL \url{https://aclanthology.org/2020.acl-main.500}.

\bibitem[Wang et~al.(2021)Wang, Jatowt, and Yoshikawa]{wang2021archivalqa}
Wang, J., Jatowt, A., and Yoshikawa, M.
\newblock Archivalqa: A large-scale benchmark dataset for open domain question
  answering over archival news collections, 2021.

\bibitem[Zhang \& Choi(2021)Zhang and Choi]{zhang-choi-2021-situatedqa}
Zhang, M. and Choi, E.
\newblock {S}ituated{QA}: Incorporating extra-linguistic contexts into {QA}.
\newblock In \emph{Proceedings of the 2021 Conference on Empirical Methods in
  Natural Language Processing}, pp.\  7371--7387, Online and Punta Cana,
  Dominican Republic, November 2021. Association for Computational Linguistics.
\newblock URL \url{https://aclanthology.org/2021.emnlp-main.586}.

\end{thebibliography}
\bibliographystyle{icml2022}

\newpage
\appendix
\onecolumn

\section{StreamingQA Dataset}
\label{sec:app_dataset}

\begin{table*}[t]
    \caption{Related datasets overview. Abbreviations: Wri-other=filtered from other datasets; KC=knowledge corpus; QD=question date; APD=article publication date.}
    \label{tab:related_datasets}
    \centering
\tiny
\setlength{\tabcolsep}{2pt}
    \begin{tabular}{p{0.18\linewidth}llllllllll}
\toprule
Dataset                                                                      
 &             
 & Knowledge Corpus (KC) & Closed/Open  & KC Size   & Train & Valid & Eval       & QD  & APD                                     & Temporal Qs \\
\midrule
\textbf{StreamingQA (this work)}                                             
 & Wri+Gen     
 & News (WMT07-20)       & OB           & 11M / 47.6M & 100k  & 10k   & 28k + 8.8k & Yes & Yes                                     & Yes         \\
\midrule
TempLama \cite{dhingra2021timeaware}                                         
 & Templ./Cloze
 & News (CustomNews)     & CB           & --        & 10k   & 5k    & 35k        & X   & Yes                                     & Yes         \\
ArchivalQA \cite{wang2021archivalqa}                                         
 & Gen         
 & News (NYT86-07)       & OB           & 1.8M      & 850k  & 100k  & 100k       & X   & Yes (transform relative time using APD) & Yes         \\
SituatedQA-temporal \cite{zhang-choi-2021-situatedqa}                        
 & Wri-other   
 & Wikipedia             & OB           & 6M / 21M  & 6k    & 3.4k  & 2.8k       & Yes & X                                       & Yes         \\
Time-Sensitive QA \cite{chen2021dataset}                                     
 & Templ./Wri  
 & Wikipedia             & OB           & 6M / 21M  & 29k   & 6.1k  & 6.1k       & X   & X                                       & Yes         \\
\midrule
Nartural Questions \cite{47761-NaturalQuestions,lee-etal-2019-latent-NQ-OPEN}
 & Wri         
 & Wikipedia             & OB           & 6M / 21M  & 79.2k & 8.8k  & 3.6k       & X   & X                                       & X           \\
PAQ \cite{lewis2021paq}                                                      
 & Gen         
 & Wikipedia             & OB           & 6M / 21M  & 65M   & X     & X          & X   & X                                       & X           \\
\midrule
CronQuestions \cite{saxena-etal-2021-question}                               
 & Templ.      
 & KG                    & KG           &           & 350k  & 30k   & 30k        & X   & KG with temporal information            & Yes         \\
TempQuestions \cite{10.1145/3184558.3191536}                                 
 & Wri-other   
 & KG                    & KG           &           &       &       & 1.2k       &     &                                         & Yes         \\
TimeQuestions \cite{timequestions2021}                                       
 & Wri-other   
 & KG                    & KG           &           & 9.7k  & 3.2k  & 3.2k       & X   & KG with temporal information            & Yes         \\
TORQUE \cite{ning-etal-2020-torque}                                          
 & Wri         
 & 3.2k short passages   & Known-source &           & 24.5k & 1.5k  & 4.6k       & X   & X                                       & Yes     \\
\bottomrule
    \end{tabular}
\end{table*}

\subsection{Comparison of StreamingQA with other related datasets}
\label{sec:app_datasets_table}
In Table~\ref{tab:related_datasets}, we present a comparison of StreamingQA with other related datasets.

\subsection{Answer types in evaluation set}
\label{sec:app_dataset_stats}
\begin{figure}
    \centering
    \includegraphics[width=0.47\linewidth]{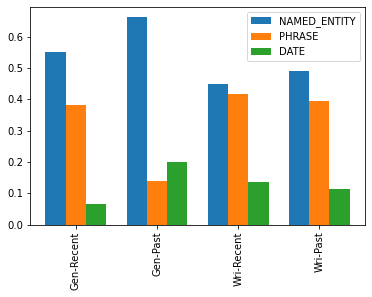}
    \caption{Answer type proportions in the \streamingqa\ evaluation sets.}
    \label{fig:stats_answer_types}
\end{figure}

We provide further detail on the answer types in our evaluation set in Figure~\ref{fig:stats_answer_types}.

\begin{figure}
    \centering
    \includegraphics[width=0.4\linewidth]{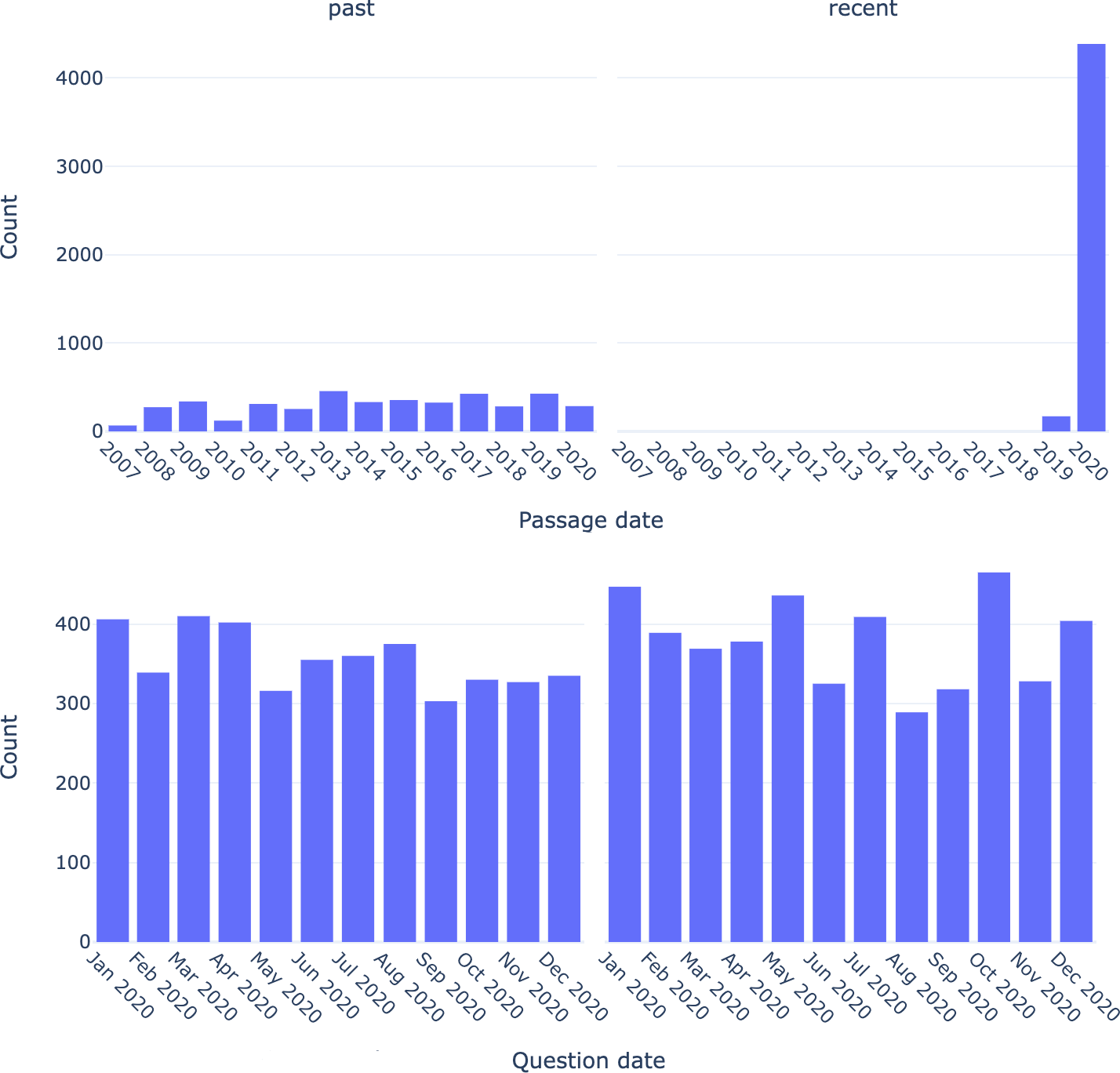}
    \caption{Question and article publication date distribution for Eval-Written. Eval-Generated is similar.}
    \label{fig:qd_apd_dist}
\end{figure}

\subsection{Examples of questions with high- and low-frequency named entities}
\label{sec:ap_dataset_examples}

\begin{table}
    \caption{Examples of questions with high and low frequency named entities.}
    \label{tab:freq_examples}
    \centering
    \small
    \begin{tabular}{p{0.9\linewidth}}
    \toprule
    \textit{Questions with high frequency named entities} \\
    When will Google have its annual I/O conference? \\
    What is the name of the Greek Prime Minister's residence? \\
         Which former New York City Mayor is developing mobile apps to help New York state trace coronavirus cases? \\
    \midrule
    \textit{Questions with low frequency named entities} \\
     What is the name of the Managing Director and Chief Executive of Latitude Financial Services?\\
     Which actress plays Vanessa Woodfield?\\
      Why must John Momis step down?\\
    \bottomrule
    \end{tabular}
\end{table}

See examples in Table~\ref{tab:freq_examples} of question with high and low frequency named entities.

\subsection{Automatic filtering of trivial and/or low-quality generated questions}
\label{sec:app_automatic_filters}
In order to remove trivial and/or low-quality questions, we apply the following filters: (i) remove questions that contain their answer as a sub-span; (ii) few-shot prompt a large LM for QA and ensure it generated the original target answer given the evidence document and the generated question for named entity or date answers exactly, and for phrases with 40\% words overlap; (iii) additionally we perform Google Search via the Google Search API\footnote{\url{https://developers.google.com/custom-search}} with the question text and evidence publication date as a query, and keep only questions for which the answer is present in the top 10 search results; and (iv) for phrase-answer questions, we only keep questions that contain a named entity in the question, hence eliminating questions that are too generic.

\subsection{Adding time specification to past generated questions}
\label{sec:time_spec}
For all questions we include an absolute or a relative time specification, choosing randomly between the two. For absolute time specification we include the month and the year of the article's publication date, for example, "in May 2017". For relative, we compute the difference between the question date and the article publication date, and include "" ($<7$ days), "a week ago" ($<1$ days), "N weeks ago" ($<=8$ weeks), "N months ago" ($<2$ years), or "N years ago".

\clearpage

\subsection{Dataset Toxicity Filtering}
\label{ap:toxicity}
We have adapted the Perspective API definitions\footnote{\url{https://developers.perspectiveapi.com/s/about-the-api-attributes-and-languages}} of the classifiers from the Perspective API to better fit our domain for our manual annotation to determine the filtering thresholds.

\begin{itemize}
    \item IDENTITY\_ATTACK:
Negative, discriminatory, stereotyping, or hateful against a group of people based on criteria including (but not limited to) race or ethnicity, religion, gender, nationality or citizenship, disability, age, or sexual orientation. As well as the above we also consider a QA if it: Unnecessarily strengthens negative, discriminatory, stereotyping, or hateful  representations of minorities. The implication of the QA could be viewed as potentially negative, discriminatory, stereotyping, or hateful and the veracity is questionable - i.e. we would need to fact-check.
    \item INSULT:
Inflammatory, insulting, or negative language towards a person or a group of people. Not necessarily identity specific.
    \item PROFANITY:
Swear words, curse words, or other obscene or profane language.
    \item SEVERE\_TOXICITY:
A very hateful, aggressive, disrespectful comment or otherwise very likely to make a user leave a discussion or give up on sharing their perspective. This attribute is much less sensitive to more mild forms of toxicity, such as comments that include positive uses of curse words.
    \item SEXUALLY\_EXPLICIT:
Contains references to lewd content. References sexual acts or body parts that are unnecessarily  graphic or detailed.
    \item THREAT:
Language that is threatening or encouraging violence or harm, including self-harm. Language that is unnecessarily graphic or detailed when reporting about a violent incident.
\end{itemize}

\section{Experiments and Results}

\subsection{Closed-book}
\label{sec:app_closedbook}

\begin{figure}
    \centering
    \includegraphics[width=0.47\linewidth]{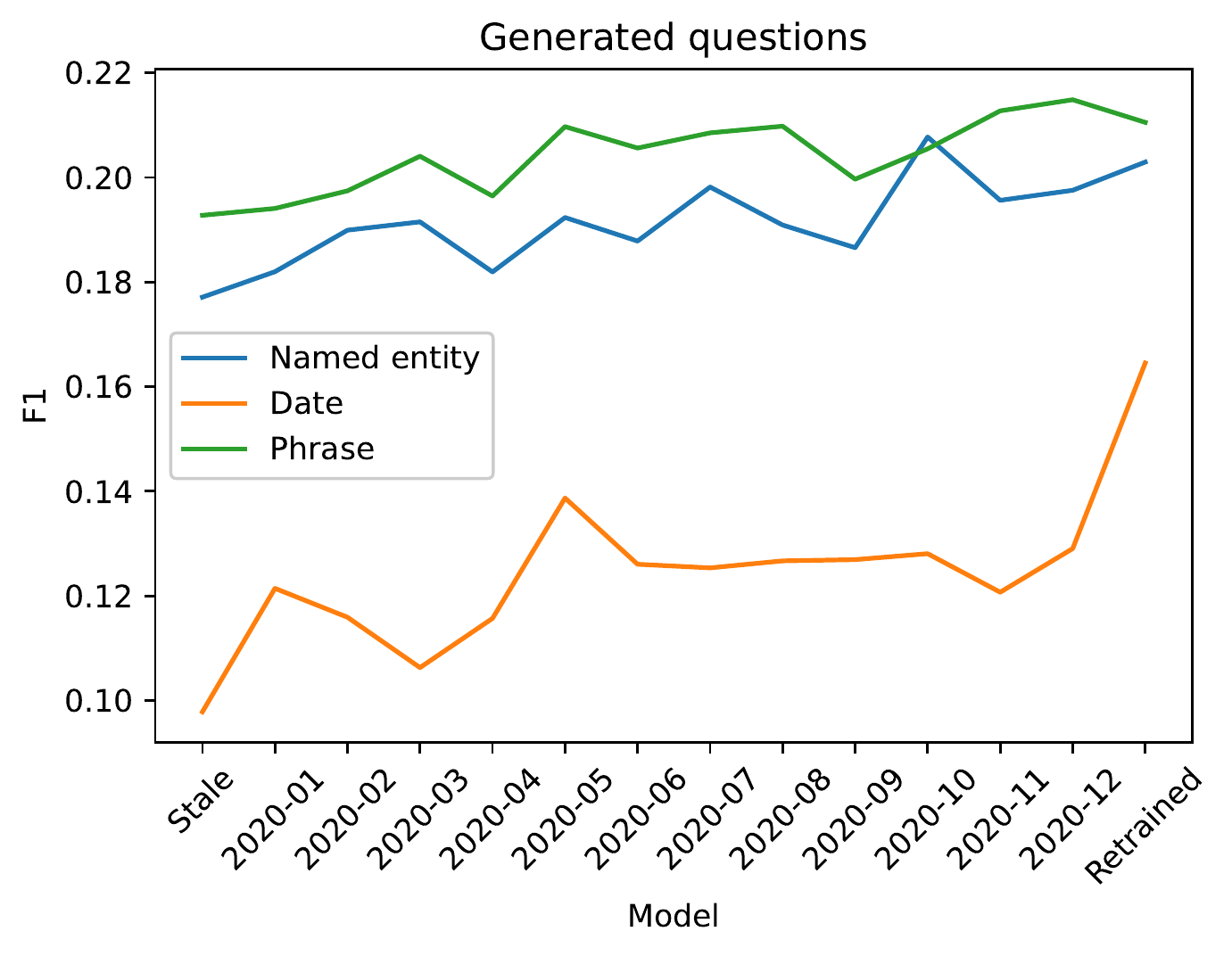}
    \includegraphics[width=0.47\linewidth]{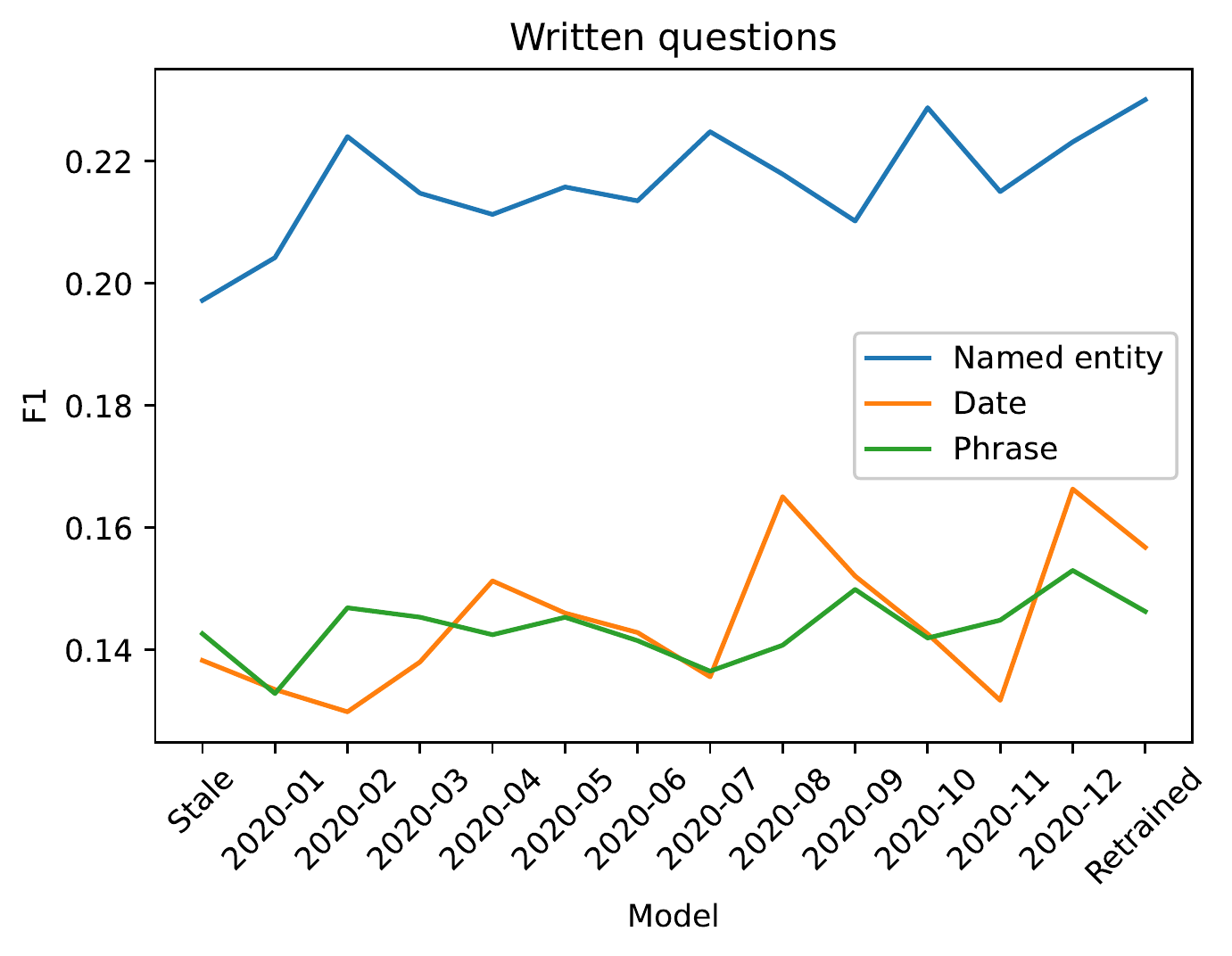}    
    \caption{F1 score on the whole dataset of models fine-tuned on data until different cut-off dates, by answer type.}
    \label{fig:closed_book_overall_by_answer_type}
\end{figure}

\begin{figure}
    \centering
    \includegraphics[width=0.47\linewidth]{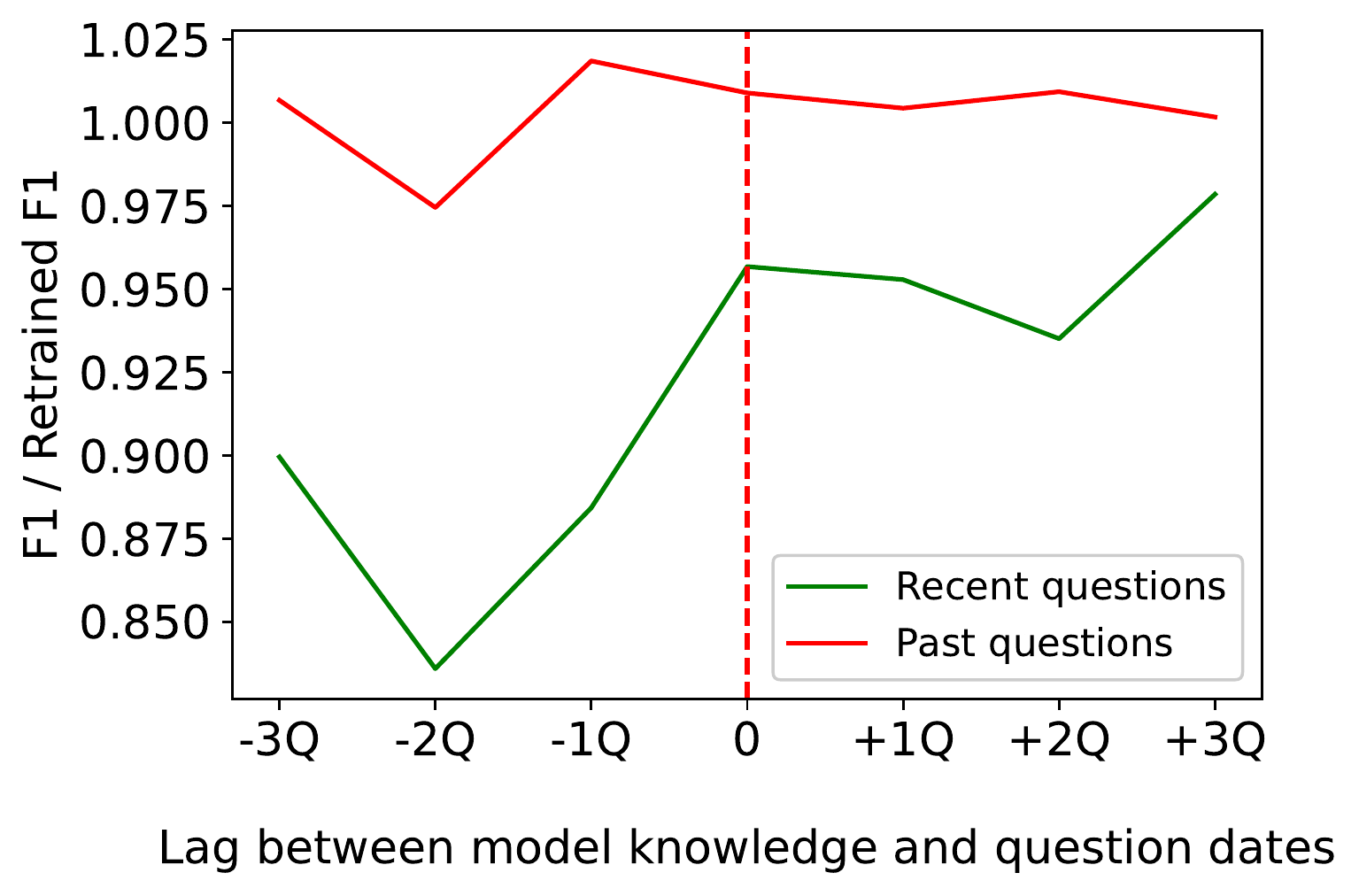}
    \caption{The effect of temporal lag between the final training month of \cbFT\ and question dates for written questions, relative to \cbRetrained.}
    \label{fig:closed_book_temporal_lag_written}
\end{figure}
\begin{figure}
    \centering
    \includegraphics[width=0.49\linewidth]{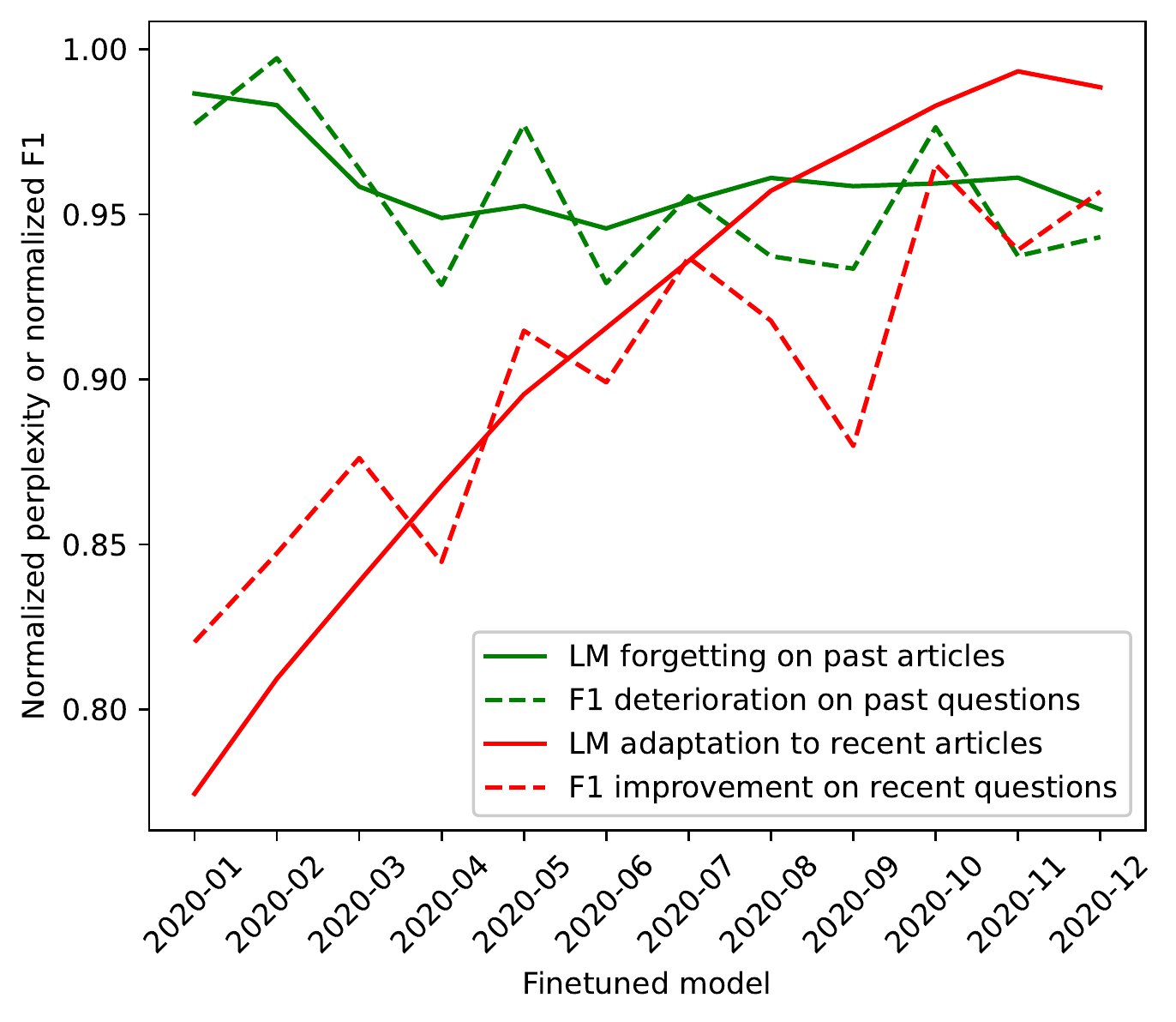}
    \caption{Relationship between adaptation and forgetting of LM (solid) and \cb\ QA models (dashed).
    Red lines show fine-tuned vs stale adaptation/improvements to recent articles on the recent subset.
    Green lines show fine-tuned vs retrained forgetting of past articles on the past subset. LM forgetting is expressed as \txlStale\ perplexity / \txlIterative\ perplexity and F1 deterioration as \cbFT\ F1\ / \cbStale\ F1, whereas LM adaptation is expressed as \txlRetrained\ perplexity / \txlIterative\ perplexity and F1 improvement as \cbFT\ F1 / \cbRetrained\ F1.}
    \label{fig:closed_book_adaptation_forgetting}
\end{figure}

We present
the F1 scores broken down by answer type in Figure~\ref{fig:closed_book_overall_by_answer_type},
and the effect of temporal lag between model knowledge and question dates for written questions in Figure~\ref{fig:closed_book_temporal_lag_written}.

\paragraph{Perplexity vs closed-book QA performance.} An interesting point of comparison is between closed-book QA performance and the perplexity of the underlying LM on test documents. As \txlIterative\ is fine-tuned on more months, we expect its perplexity on evidence documents of the recent subset
to reduce, while its perplexity on evidence documents of the past subset
to either stay the same (in the optimal scenario) or increase, if the model forgets. 
Figure~\ref{fig:closed_book_adaptation_forgetting} shows these two effects.

\subsection{Open-book}
\label{sec:app_openbook}

\begin{figure}
    \centering
    \includegraphics[width=0.47\linewidth]{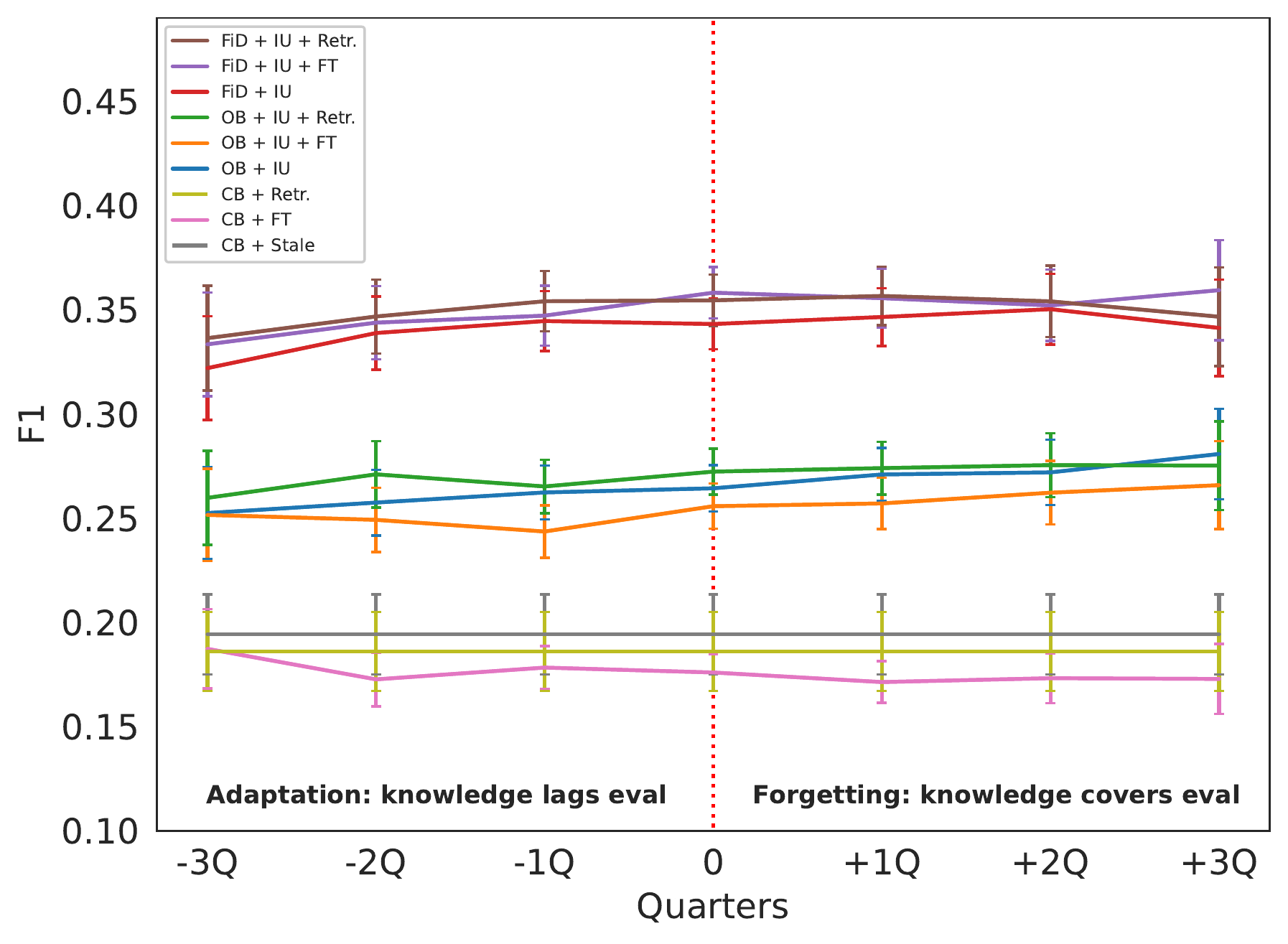}
    \caption{Adaptation and forgetting on past written questions: Open-book versus closed-book.}
    \label{fig:adaptation_past_wri}
\end{figure}
\begin{figure}
    \centering
    \includegraphics[width=0.47\linewidth]{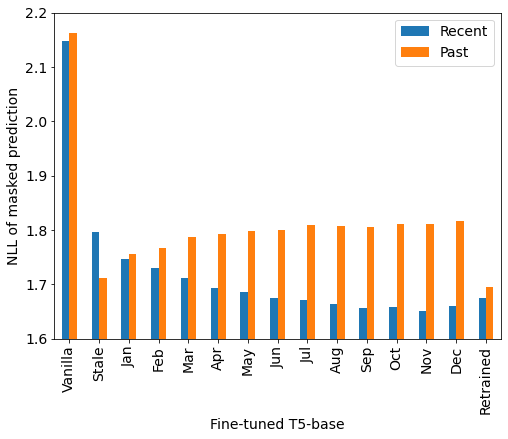}
    \caption{Negative log likelihood of masked span prediction for evaluation documents of the T5. We show the vanilla T5, T5 fine-tuned on WMT up to 2019 ("Stale"), monthly fine-tuned, and trained on all WMT up to 2020. We see forgetting on the past subset, and a slight recency bias on the recent subset, compared to retraining.}
    \label{fig:t5_base_docs}
\end{figure}

We show adaptation and forgetting on past written questions in Figure~\ref{fig:adaptation_past_wri}, metrics for all subsets in Table~\ref{tab:adaptation_numbers}, and
the masked span prediction performance for evaluation documents of the T5 model in Figure~\ref{fig:t5_base_docs}.

\begin{table}[t]
    \caption{Adaptation and forgetting F1 scores (with 95\% confidence intervals).}
    \label{tab:adaptation_numbers}
    \centering
    \tiny
    \begin{tabular}{lccccccc}
    \toprule
    Model & -3Q & -2Q & -1Q & 0Q & +1Q & +2Q & +3Q \\
    \midrule
    \textbf{Eval-Generated, Recent} \\
CB + Stale         &  $0.1845 \pm 0.0096$  &  $0.1845 \pm 0.0096$  &  $0.1845 \pm 0.0096$  &  $0.1845 \pm 0.0096$  &  $0.1845 \pm 0.0096$  &  $0.1845 \pm 0.0096$  &  $0.1845 \pm 0.0096$  \\
CB + FT            &  $0.1968 \pm 0.0098$  &  $0.2007 \pm 0.0072$  &  $0.2085 \pm 0.0059$  &  $0.2270 \pm 0.0054$  &  $0.2222 \pm 0.0062$  &  $0.2194 \pm 0.0075$  &  $0.2172 \pm 0.0109$  \\
CB + Retr.         &  $0.2389 \pm 0.0109$  &  $0.2389 \pm 0.0109$  &  $0.2389 \pm 0.0109$  &  $0.2389 \pm 0.0109$  &  $0.2389 \pm 0.0109$  &  $0.2389 \pm 0.0109$  &  $0.2389 \pm 0.0109$  \\
OB + IU            &  $0.1970 \pm 0.0103$  &  $0.2099 \pm 0.0076$  &  $0.2459 \pm 0.0066$  &  $0.3372 \pm 0.0065$  &  $0.3435 \pm 0.0076$  &  $0.3360 \pm 0.0092$  &  $0.3287 \pm 0.0132$  \\
OB + IU + FT       &  $0.1973 \pm 0.0103$  &  $0.2059 \pm 0.0076$  &  $0.2481 \pm 0.0067$  &  $0.3351 \pm 0.0065$  &  $0.3409 \pm 0.0076$  &  $0.3348 \pm 0.0092$  &  $0.3172 \pm 0.0130$  \\
OB + IU + Retr.    &  $0.2301 \pm 0.0111$  &  $0.2434 \pm 0.0081$  &  $0.2734 \pm 0.0069$  &  $0.3672 \pm 0.0067$  &  $0.3696 \pm 0.0078$  &  $0.3599 \pm 0.0094$  &  $0.3532 \pm 0.0135$  \\
FiD + IU           &  $0.2471 \pm 0.0113$  &  $0.2678 \pm 0.0084$  &  $0.3413 \pm 0.0075$  &  $0.5653 \pm 0.0070$  &  $0.5649 \pm 0.0081$  &  $0.5585 \pm 0.0099$  &  $0.5533 \pm 0.0143$  \\
FiD + IU + FT      &  $0.2487 \pm 0.0115$  &  $0.2703 \pm 0.0085$  &  $0.3440 \pm 0.0075$  &  $0.5817 \pm 0.0069$  &  $0.5834 \pm 0.0081$  &  $0.5734 \pm 0.0098$  &  $0.5636 \pm 0.0142$  \\
FiD + IU + Retr.   &  $0.2520 \pm 0.0114$  &  $0.2743 \pm 0.0085$  &  $0.3444 \pm 0.0075$  &  $0.5745 \pm 0.0070$  &  $0.5701 \pm 0.0081$  &  $0.5632 \pm 0.0098$  &  $0.5618 \pm 0.0143$  \\
    \midrule
    \textbf{Eval-Generated, Past} \\
CB + Stale         &  $0.1541 \pm 0.0105$  &  $0.1541 \pm 0.0105$  &  $0.1541 \pm 0.0105$  &  $0.1541 \pm 0.0105$  &  $0.1541 \pm 0.0105$  &  $0.1541 \pm 0.0105$  &  $0.1541 \pm 0.0105$  \\
CB + FT            &  $0.1523 \pm 0.0101$  &  $0.1452 \pm 0.0070$  &  $0.1488 \pm 0.0058$  &  $0.1515 \pm 0.0052$  &  $0.1461 \pm 0.0059$  &  $0.1499 \pm 0.0073$  &  $0.1442 \pm 0.0103$  \\
CB + Retr.         &  $0.1517 \pm 0.0103$  &  $0.1517 \pm 0.0103$  &  $0.1517 \pm 0.0103$  &  $0.1517 \pm 0.0103$  &  $0.1517 \pm 0.0103$  &  $0.1517 \pm 0.0103$  &  $0.1517 \pm 0.0103$  \\
OB + IU            &  $0.2757 \pm 0.0138$  &  $0.2857 \pm 0.0100$  &  $0.2970 \pm 0.0083$  &  $0.3012 \pm 0.0073$  &  $0.3050 \pm 0.0086$  &  $0.3066 \pm 0.0106$  &  $0.3089 \pm 0.0151$  \\
OB + IU + FT       &  $0.2735 \pm 0.0137$  &  $0.2750 \pm 0.0098$  &  $0.2839 \pm 0.0082$  &  $0.2888 \pm 0.0072$  &  $0.2924 \pm 0.0085$  &  $0.2927 \pm 0.0105$  &  $0.2948 \pm 0.0150$  \\
OB + IU + Retr.    &  $0.2865 \pm 0.0140$  &  $0.2994 \pm 0.0102$  &  $0.3050 \pm 0.0084$  &  $0.3105 \pm 0.0074$  &  $0.3148 \pm 0.0087$  &  $0.3123 \pm 0.0107$  &  $0.3068 \pm 0.0151$  \\
FiD + IU           &  $0.4030 \pm 0.0154$  &  $0.4208 \pm 0.0112$  &  $0.4372 \pm 0.0092$  &  $0.4606 \pm 0.0081$  &  $0.4638 \pm 0.0095$  &  $0.4661 \pm 0.0117$  &  $0.4665 \pm 0.0166$  \\
FiD + IU + FT      &  $0.4125 \pm 0.0156$  &  $0.4283 \pm 0.0112$  &  $0.4481 \pm 0.0093$  &  $0.4721 \pm 0.0082$  &  $0.4754 \pm 0.0095$  &  $0.4768 \pm 0.0117$  &  $0.4686 \pm 0.0168$  \\
FiD + IU + Retr.   &  $0.4245 \pm 0.0157$  &  $0.4402 \pm 0.0113$  &  $0.4561 \pm 0.0093$  &  $0.4775 \pm 0.0082$  &  $0.4788 \pm 0.0095$  &  $0.4813 \pm 0.0117$  &  $0.4772 \pm 0.0167$  \\
    \midrule
    \textbf{Eval-Written, Recent} \\
CB + Stale         &  $0.1685 \pm 0.0162$  &  $0.1685 \pm 0.0162$  &  $0.1685 \pm 0.0162$  &  $0.1685 \pm 0.0162$  &  $0.1685 \pm 0.0162$  &  $0.1685 \pm 0.0162$  &  $0.1685 \pm 0.0162$  \\
CB + FT            &  $0.1814 \pm 0.0166$  &  $0.1721 \pm 0.0118$  &  $0.1771 \pm 0.0099$  &  $0.1909 \pm 0.0088$  &  $0.1886 \pm 0.0104$  &  $0.1802 \pm 0.0120$  &  $0.1932 \pm 0.0173$  \\
CB + Retr.         &  $0.2017 \pm 0.0176$  &  $0.2017 \pm 0.0176$  &  $0.2017 \pm 0.0176$  &  $0.2017 \pm 0.0176$  &  $0.2017 \pm 0.0176$  &  $0.2017 \pm 0.0176$  &  $0.2017 \pm 0.0176$  \\
OB + IU            &  $0.1778 \pm 0.0170$  &  $0.1752 \pm 0.0123$  &  $0.1892 \pm 0.0105$  &  $0.2661 \pm 0.0107$  &  $0.2692 \pm 0.0125$  &  $0.2777 \pm 0.0151$  &  $0.3032 \pm 0.0219$  \\
OB + IU + FT       &  $0.1739 \pm 0.0168$  &  $0.1664 \pm 0.0120$  &  $0.1860 \pm 0.0105$  &  $0.2678 \pm 0.0107$  &  $0.2659 \pm 0.0125$  &  $0.2786 \pm 0.0152$  &  $0.2911 \pm 0.0217$  \\
OB + IU + Retr.    &  $0.2008 \pm 0.0176$  &  $0.2050 \pm 0.0132$  &  $0.2134 \pm 0.0111$  &  $0.2887 \pm 0.0110$  &  $0.2905 \pm 0.0129$  &  $0.2973 \pm 0.0156$  &  $0.3131 \pm 0.0223$  \\
FiD + IU           &  $0.1843 \pm 0.0176$  &  $0.1894 \pm 0.0131$  &  $0.2288 \pm 0.0118$  &  $0.3816 \pm 0.0121$  &  $0.3874 \pm 0.0142$  &  $0.3920 \pm 0.0170$  &  $0.4118 \pm 0.0237$  \\
FiD + IU + FT      &  $0.1768 \pm 0.0174$  &  $0.1957 \pm 0.0133$  &  $0.2356 \pm 0.0120$  &  $0.4086 \pm 0.0122$  &  $0.4078 \pm 0.0143$  &  $0.4096 \pm 0.0171$  &  $0.4345 \pm 0.0241$  \\
FiD + IU + Retr.   &  $0.1848 \pm 0.0178$  &  $0.2031 \pm 0.0135$  &  $0.2402 \pm 0.0120$  &  $0.3971 \pm 0.0121$  &  $0.3966 \pm 0.0141$  &  $0.4031 \pm 0.0170$  &  $0.4242 \pm 0.0239$  \\
    \midrule
    \textbf{Eval-Written, Past} \\
CB + Stale         &  $0.1943 \pm 0.0193$  &  $0.1943 \pm 0.0193$  &  $0.1943 \pm 0.0193$  &  $0.1943 \pm 0.0193$  &  $0.1943 \pm 0.0193$  &  $0.1943 \pm 0.0193$  &  $0.1943 \pm 0.0193$  \\
CB + FT            &  $0.1873 \pm 0.0190$  &  $0.1727 \pm 0.0129$  &  $0.1784 \pm 0.0104$  &  $0.1760 \pm 0.0089$  &  $0.1714 \pm 0.0100$  &  $0.1733 \pm 0.0120$  &  $0.1729 \pm 0.0167$  \\
CB + Retr.         &  $0.1861 \pm 0.0190$  &  $0.1861 \pm 0.0190$  &  $0.1861 \pm 0.0190$  &  $0.1861 \pm 0.0190$  &  $0.1861 \pm 0.0190$  &  $0.1861 \pm 0.0190$  &  $0.1861 \pm 0.0190$  \\
OB + IU            &  $0.2525 \pm 0.0222$  &  $0.2575 \pm 0.0158$  &  $0.2624 \pm 0.0130$  &  $0.2644 \pm 0.0111$  &  $0.2710 \pm 0.0128$  &  $0.2720 \pm 0.0156$  &  $0.2808 \pm 0.0217$  \\
OB + IU + FT       &  $0.2516 \pm 0.0222$  &  $0.2492 \pm 0.0154$  &  $0.2436 \pm 0.0126$  &  $0.2558 \pm 0.0109$  &  $0.2571 \pm 0.0124$  &  $0.2623 \pm 0.0152$  &  $0.2659 \pm 0.0211$  \\
OB + IU + Retr.    &  $0.2598 \pm 0.0226$  &  $0.2711 \pm 0.0159$  &  $0.2652 \pm 0.0128$  &  $0.2724 \pm 0.0110$  &  $0.2740 \pm 0.0127$  &  $0.2755 \pm 0.0154$  &  $0.2753 \pm 0.0213$  \\
FiD + IU           &  $0.3220 \pm 0.0249$  &  $0.3389 \pm 0.0175$  &  $0.3446 \pm 0.0143$  &  $0.3432 \pm 0.0122$  &  $0.3465 \pm 0.0139$  &  $0.3503 \pm 0.0169$  &  $0.3413 \pm 0.0231$  \\
FiD + IU + FT      &  $0.3334 \pm 0.0249$  &  $0.3438 \pm 0.0175$  &  $0.3472 \pm 0.0144$  &  $0.3582 \pm 0.0123$  &  $0.3556 \pm 0.0140$  &  $0.3521 \pm 0.0171$  &  $0.3594 \pm 0.0240$  \\
FiD + IU + Retr.   &  $0.3364 \pm 0.0251$  &  $0.3468 \pm 0.0177$  &  $0.3541 \pm 0.0145$  &  $0.3545 \pm 0.0123$  &  $0.3566 \pm 0.0141$  &  $0.3540 \pm 0.0172$  &  $0.3467 \pm 0.0237$  \\
\bottomrule
    \end{tabular}
\end{table}

\subsection{Temporal accuracy of DPR trained on news}\label{sec:ap_tempretr}
\begin{figure}
    \centering
    \includegraphics[width=0.6\linewidth]{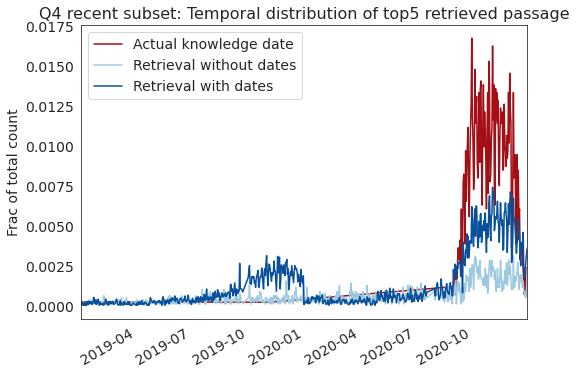}
    \caption{
    Temporal distribution of gold and retrieved passages of the recent questions for Q4'2020. DPR with timestamp matches the temporal distribution of the gold passages much closer. Interestingly, the model seems to get somewhat confused about the year - the second lower spike is in Q4'2019.
    }
    \label{fig:dpr_temporal_dist}
\end{figure}

Figure \ref{fig:dpr_temporal_dist} shows temporal distribution of gold and retrieved passages for recent questions for Q4'2020: DPR with timestamp matches the temporal distribution of the gold passages much closer. Interestingly, the model seems to get somewhat confused about the year: the second lower spike is in Q4'2019.

\subsection{One-step Streaming and Static QA Benchmarks}
\label{sec:app_static}

\begin{figure}
    \centering
    \includegraphics[width=0.65\linewidth]{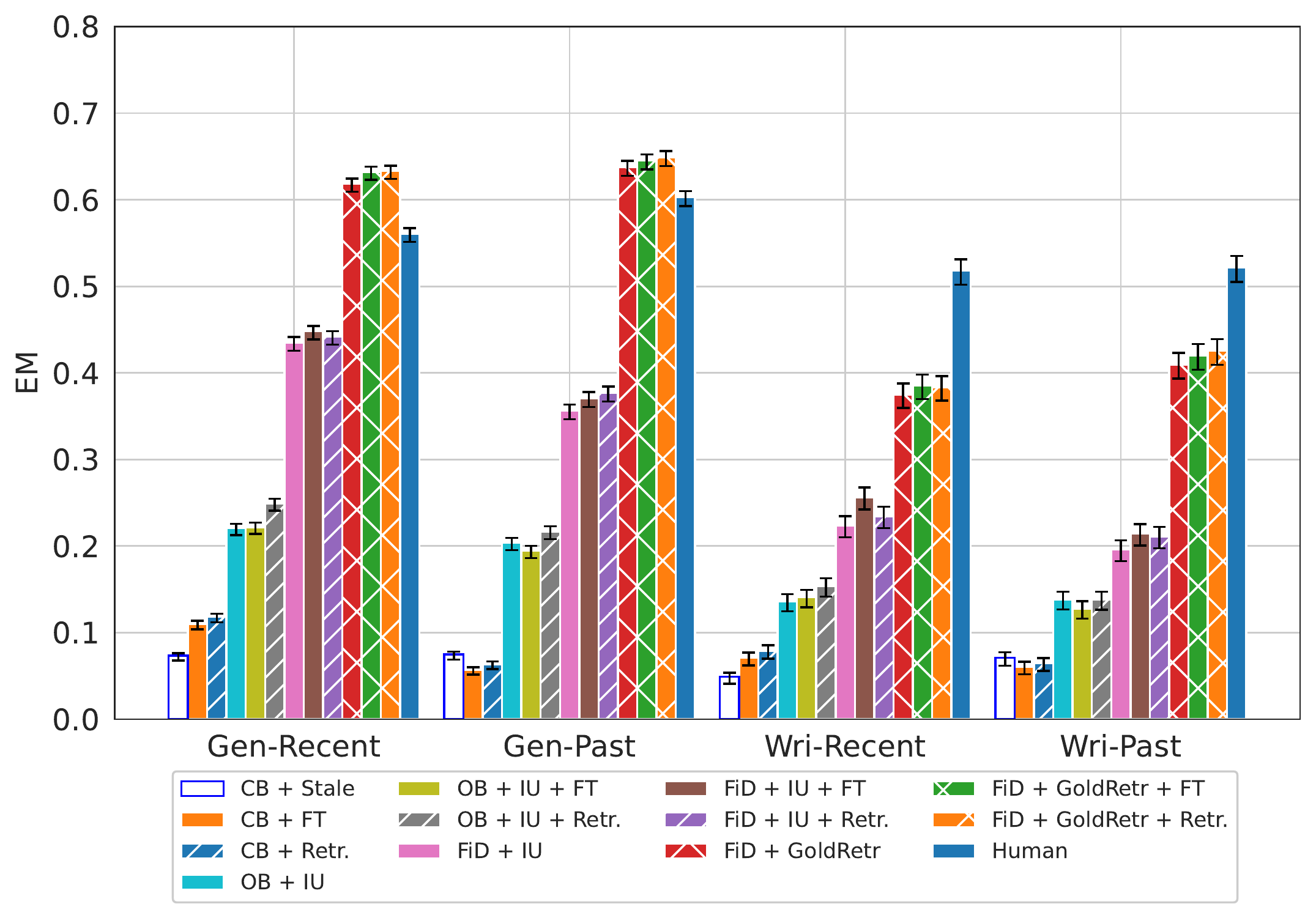}
    \caption{Static setup with all questions. Solid filled bars are models that had 2020 knowledge added incrementally.}
    \label{fig:static_all_em}
\end{figure}

We provide the EM in Figure~\ref{fig:static_all_em} and all the metrics in Table~\ref{tab:static}.

\begin{table}[t]
    \caption{Static setup.}
    \label{tab:static}
    \centering
    \tiny
    \begin{tabular}{lcccc}
    \toprule
    Model &  \multicolumn{2}{c}{Recent} &  \multicolumn{2}{c}{Past} \\
          &  EM  &  F1 & EM  &  F1 \\
    \midrule
    \textbf{Generated} \\
CB + Stale              & 0.0736 & 0.1822 & 0.0750 & 0.1577 \\
CB + FT                 & 0.1102 & 0.2287 & 0.0573 & 0.1487 \\
CB + Retr.              & 0.1184 & 0.2390 & 0.0637 & 0.1503 \\
OB + IU                 & 0.2206 & 0.3393 & 0.2039 & 0.3020 \\
OB + IU + FT            & 0.2219 & 0.3383 & 0.1948 & 0.2879 \\
OB + IU + Retr.         & 0.2493 & 0.3685 & 0.2169 & 0.3117 \\
FiD + IU                & 0.4350 & 0.5616 & 0.3565 & 0.4582 \\
FiD + IU + FT           & 0.4480 & 0.5776 & 0.3707 & 0.4697 \\
FiD + IU + Retr.        & 0.4419 & 0.5697 & 0.3771 & 0.4743 \\
FiD + GoldRetr          & 0.6184 & 0.7273 & 0.6378 & 0.7205 \\
FiD + GoldRetr + FT     & 0.6320 & 0.7326 & 0.6452 & 0.7237 \\
FiD + GoldRetr + Retr.  & 0.6332 & 0.7339 & 0.6491 & 0.7267 \\
Human                   & 0.5608 & 0.7744 & 0.6027 & 0.7809 \\
    \midrule
    \textbf{Written} \\
CB + Stale              & 0.0489 & 0.1528 & 0.0710 & 0.1837 \\
CB + FT                 & 0.0710 & 0.1944 & 0.0608 & 0.1804 \\
CB + Retr.              & 0.0792 & 0.1987 & 0.0648 & 0.1744 \\
OB + IU                 & 0.1360 & 0.2689 & 0.1385 & 0.2648 \\
OB + IU + FT            & 0.1409 & 0.2681 & 0.1277 & 0.2541 \\
OB + IU + Retr.         & 0.1537 & 0.2888 & 0.1383 & 0.2700 \\
FiD + IU                & 0.2238 & 0.3803 & 0.1961 & 0.3443 \\
FiD + IU + FT           & 0.2564 & 0.4135 & 0.2145 & 0.3552 \\
FiD + IU + Retr.        & 0.2345 & 0.3969 & 0.2112 & 0.3532 \\
FiD + GoldRetr          & 0.3751 & 0.5677 & 0.4097 & 0.5985 \\
FiD + GoldRetr + FT     & 0.3853 & 0.5721 & 0.4201 & 0.6003 \\
FiD + GoldRetr + Retr.  & 0.3835 & 0.5697 & 0.4258 & 0.6040 \\
Human                   & 0.5180 & 0.7445 & 0.5216 & 0.7405 \\
\bottomrule
    \end{tabular}
\end{table}

\end{document}